\documentclass[runningheads,a4paper]{llncs}
\usepackage{times}
\usepackage{graphicx}
\usepackage{helvet}
\usepackage{courier}  
\usepackage{helvet} 
\usepackage{latexsym}
\usepackage{amsmath} 
\usepackage{amssymb}
\usepackage{xspace}
\usepackage{algorithm}
\usepackage{algorithmic}
\usepackage{verbatim}
\usepackage{float} 
\usepackage{url}
\usepackage{rotating}
\usepackage[sectionbib,sort&compress,numbers,square]{natbib}
\usepackage{tikz}
\usetikzlibrary{arrows,snakes,backgrounds,automata,shapes,patterns}
\newenvironment{Proof}{\textsc{Proof.\ }\normalfont}{\qed}

\long\def\eatpar#1{%
\ifx#1\par                      % se il token e' \par
\let\nextmove=\eatpar           % rimetti \eatpar in coda
\else
\let\nextmove=#1%               altrimenti, rimetti il token in coda
\fi
\nextmove%                      il token o \eatpar viene rimesso in coda
}

\def\qed{\hfill{\qedboxempty}      % qed with empty box
  \ifdim\lastskip<\medskipamount \removelastskip\penalty55\medskip\fi}

\def\qedboxempty{\vbox{\hrule\hbox{\vrule\kern3pt
                 \vbox{\kern3pt\kern3pt}\kern3pt\vrule}\hrule}}

\def\qedfull{\hfill{\qedboxfull}   % qed with full box
  \ifdim\lastskip<\medskipamount \removelastskip\penalty55\medskip\fi}

\def\qedboxfull{\vrule height 4pt width 4pt depth 0pt}

\newcommand{{\incolumn}}[1]{\begin{tabular}[c]{c} #1 \end{tabular}}
\newcommand{{\incolumnmath}}[1]{\begin{array}[c]{c} #1 \end{array}}
\begingroup
\catcode`\~=11
\gdef\urltilde{\lower 0.6ex\hbox{~}}
\endgroup

\newcommand{\A}{\mathcal{A}} \newcommand{\B}{\mathcal{B}}
\newcommand{\C}{\mathcal{C}} 
\newcommand{\E}{\mathcal{E}} 
 \renewcommand{\H}{\mathcal{H}}

\newcommand{\M}{\mathcal{M}} 
 
 \newcommand{\R}{\mathcal{R}}
\renewcommand{\S}{\mathcal{S}} \newcommand{\T}{\mathcal{T}}
\newcommand{\U}{\mathcal{U}}

\newcommand{\defterm}[1]{\mbox{\underline{\it\smash{#1}\vphantom{\lower.1ex\hbox
{x}}}}}

\newcommand{\lora}{\longrightarrow}

\newcommand{\goto}[1]{\stackrel{#1}{\lora}}

\newcommand{\set}[1]{\{#1\}}                      % set

\newcommand{\card}[1]{|{#1}|}                     % cardinality of a set

\newcommand{\tup}[1]{\langle #1\rangle}            % tuple

\newcommand{\name}[1]{\ensuremath{\operatorname{\textit{#1}}}}

\newcommand{\beh}{\name{beh}}

\newcommand{\last}{\name{last}}

\newcommand{\projact}[1]{[#1]}

\newcommand{\comp}{\mathit{comp}}

\newcommand{\myi}{\emph{(i)}\xspace}
\newcommand{\myii}{\emph{(ii)}\xspace}
\newcommand{\myiii}{\emph{(iii)}\xspace}
\newcommand{\myiv}{\emph{(iv)}\xspace}

\newcommand{\Nat}{{\rm I\kern-.23em N}}

\newcommand{\commentout}[1]{}

\tikzstyle{every initial by arrow}=[initial text=]
\tikzstyle{every state}=[fill=none,draw=black,text=black]
\tikzstyle{every state}=[fill=none,draw=black,text=black,inner sep=0pt,minimum
size=7mm]
\tikzstyle{every picture}=[->,>=stealth',shorten >=1pt,auto,node distance=2.5cm,
semithick]
\tikzstyle{sim}=[->,dotted]
\newcommand{\labelfig}[1]{\textcolor{blue}{\sc #1}}

\usepackage{marginnote}
\edef\marginnotetextwidth{\the\textwidth}

\newcommand{\propername}[1]{\text{\textsc{#1}}\xspace}
\newcommand{\actionfont}[1]{\text{\textsc{#1}}}

\newcommand{\aLightOn}{\actionfont{lightOn}}
\newcommand{\aLightOff}{\actionfont{lightOff}}

\newcommand{\aMusic}{\actionfont{music}}
\newcommand{\aMovie}{\actionfont{movie}}
\newcommand{\aGame}{\actionfont{game}}
\newcommand{\aRadio}{\actionfont{radio}}
\newcommand{\aStop}{\actionfont{stop}}

\newcommand{\aWeb}{\actionfont{web}}
\newcommand{\aUnplug}{\actionfont{unplug}}

\newcommand{\citea}[1]{\citeauthor{#1}~\cite{#1}}

\newcommand{\expl}{\propername{Expl}}

\newcommand{\aprxtgt}{\ensuremath{\tilde{\T}}}
\newcommand{\aprxcntrl}{\ensuremath{C_{\tilde{\T}}}}

\newcommand{\TEX}{\T_{\text{\textsc{ENT}}}}
\newcommand{\aprxtgtEX}{\aprxtgt_{\text{\textsc{ENT}}}}

\newcommand{\Sim}{\operatorname{\emph{Sim}}}

\begin{document}
\mainmatter  % start of an individual contribution

\title{Qualitative Approximate Behavior Composition}

%\titlerunning{Lecture Notes in Computer Science: Authors' Instructions}

% % the name(s) of the author(s) follow(s) next
\author{Nitin Yadav and Sebastian Sardina 
\thanks{We acknowledge the support of the Australian Research Council under grant DP120100332.}
}

%\institute{RMIT University, Melbourne, Australia.}

\renewcommand\bibname{References}
\institute{RMIT University, Melbourne, Australia.}
\maketitle

\begin{abstract}

The behavior composition problem involves automatically building a controller  that is able to realize 
%(i.e., implement) 
a desired, but unavailable, target system (e.g., a house surveillance) by suitably coordinating a set of available components
%, though partially controllable, components
(e.g., video cameras, blinds, lamps, a vacuum cleaner, phones, etc.)
%%
% Thus, a complex house surveillance system could be realized by intelligently coordinating various devices installed throughout the space, including video cameras, automatic blinds and lamps, a vacuum cleaner, phones, etc.
% %% 
Previous work has almost exclusively aimed at bringing about the desired component in its totality, which is highly unsatisfactory for unsolvable problems.
In this work, we develop an approach for \emph{approximate} behavior composition without departing from the classical setting, 
% that caters for instances admitting no complete solutions, 
thus making the problem applicable to a much wider range of cases.
%%
% This clearly poses a major limitation in that for many problem instances, if not most,  there will
% be no exact controller. For such cases, a (merely) ``no solution" outcome is extremely unsatisfactory. 
%%
Based on the notion of simulation, we characterize what a maximal controller and the ``closest'' implementable target module (optimal approximation) are, and show how these can be  computed using ATL model checking technology for a special case. 
We show the uniqueness of optimal approximations, and prove their soundness and completeness with respect to their imported controllers.
%%
%Finally, we show how to compute solutions for a special case using ATL model checking.
\end{abstract}

\section{Introduction}\label{sec:introduction}

% This work is concerned with developing a \emph{qualitative} account of behavior composition \emph{approximation} (or \emph{optimization}), able to accommodate problem instances with no (total) solutions.

%The behavior composition problem (e.g., \cite{Balbianietal:SERVICES09,LustigVardi:FOSSACS09,DeGiacomoS:IJCAI07,StroederPagnucco:IJCAI09}) involves the automatic synthesis, that is, construction, of  an (embedded) controller that is able to ``realize" (i.e., implement) a given desired, though non-existent, complex target system (e.g., a smart house entertainment system) by suitably coordinating  a collection of  partially controllable available behaviors (e.g., music devices, TVs, game stations, blinds, lights, etc.)
%%
The behavior composition problem (e.g., \cite{Balbianietal:SERVICES09,LustigVardi:FOSSACS09,DeGiacomoS:IJCAI07,StroederPagnucco:IJCAI09}) involves the automatic synthesis of a controller that is able to ``realize" (i.e., implement) a desired, though non-existent, complex target system by suitably coordinating  a collection of  partially controllable available behaviors.
A behavior here refers to the abstract operational model of a device or program, generally represented as a non-deterministic transition system.
Thus, in a smart building setting, one may look for a controller able to coordinate the execution of a set of devices installed in a house---music and movie players, game consoles, automatic blinds and lights, radios, etc.---such that it appears as if a complex entertainment system was actually being run.  
A solution to the problem is called a \emph{composition}.
%%

%A controller that is able to coordinate the execution of a set of available behaviors such that it appears that the target behavior which is actually being run, is called a \emph{composition}, a solution to the problem.
%%
%Behaviors here refer to the operational logic of devices or programs, and are generally represented as non-deterministic transition systems. 
% Behavior here refers to the abstract operational model of a device or program, and is generally represented as a non-deterministic transition system. 
%%
% The problem is indeed appealing in that with computers now present in everyday devices like mobile phones, credit cards, cars and planes, or places like homes, offices and factories, the trend is to build embedded complex systems from a collection of simple components.
% %%
% What is more, it can be easily related to other problems within several sub-areas of AI and CS, including web-service composition~\cite{Hull:IEEE-SCC05}, reactive synthesis~\cite{PnRo89}, agent-oriented programming~\cite{Shoham:AIJ93-Agent0}, robot ecologies~\cite{SaffiottiBroxvall:SOAI05}, and automated planning~\cite{GhallabNT:04-Planning}.
The composition problem is appealing to a wide range of audiences. Indeed, with computers now present in everyday devices like mobile phones, credit cards, or places like homes, offices and factories, the trend is to build embedded complex devices from a collection of simple components. 
In addition, the problem can be related to several sub-areas of AI and CS, including web-service composition~\cite{Hull:IEEE-SCC05}, reactive synthesis~\cite{PnRo89}, agent-oriented programming~\cite{Shoham:AIJ93-Agent0}, robot ecologies~\cite{SaffiottiBroxvall:SOAI05}, and automated planning~\cite{GhallabNT:04-Planning}.

% With the exception of our recent work [S8], all the literature on behaviour composition has only dealt with perfect compositions to date: the composition controller ought to guarantee the realisation of every possible target step always. This poses a major limitation in that for many, if not most, problem instances there will be no composition at all. For such cases, a (merely) ``no solution" outcome is extremely unsatisfactory. 

While the behavior composition problem has been substantially studied in an AI context lately (e.g., \cite{DeGiacomoS:IJCAI07,SardinaPDG:KR08,StroederPagnucco:IJCAI09}), previous work has exclusively aimed at the synthesis of \emph{complete} realisations of the desired target component---compositions that implement the desired component in its totality. %%
% \footnote{The work in~\cite{YadavSardina:AAMAS11} is an exception, but it is of quantitative nature and requires additional domain knowledge; see Discussion section.} 
%%
This poses a major limitation in problem instances with no (exact) compositions. For such cases, a merely ``no solution" outcome is extremely unsatisfactory. 
The need to address this shortcoming has already been noted in previous works~\cite{StroederPagnucco:IJCAI09, YadavSardina:AAMAS11}.
In this paper, we develop a qualitative account of \emph{approximate behavior composition} that caters for instances admitting no exact solutions. 
%%
% Importantly, we assume no additional domain information, and hence deal with the \emph{classical} composition setting found in the literature.

Intuitively, the overarching idea is to \emph{look for those parts of the target module that can be realized with the available modules}, and provide this as an (approximate) solution.
More precisely, given a target module, the task is to identify the \emph{closest} alternative target module that can be fully realized with the behaviors at hand---the optimal approximate target. Of course, it is expected that such alternative target will generally provide less functionalities than the original one. Indeed, some execution paths may be impossible to generate with the new target (e.g., it may no more be feasible to play video games when listening to music). Moreover, the alternative target may accommodate less ``freedom'' of choices in executions (e.g., when requesting to watch a movie, one may now need to commit to whether one will be playing a video game or listening to radio afterwards).
Nonetheless, the user can request actions as per the alternative (approximate) target and be guaranteed her requests will always be fulfilled.

Observe that in this paper we assume a setting of \emph{strict} uncertainty, in that the space of possibilities (behaviors' evolutions and target requests) is known, but the probabilities of these potential alternatives cannot be quantified~\cite{French:DT86}.
This contrasts with our previous approach~\cite{YadavSardina:AAMAS11}, which assumes all such probabilities have been specified for the domain and then looks for the ``best'' controller possible from a decision-theoretic perspective.
Consequently, our account here can be seen as the next natural extension of the ``classical'' composition framework found in the literature, in that no no additional domain information is assumed.
We shall discuss and compare this further in Section~\ref{sec:conclusion}.

% in which no such knowledge is assumed to be available. Indeed, in many scenarios, one may just not know how the user will make use of the target module or may not be able to quantify the internal non-determinism in the available behaviors.
% %%
%%
% However, in this paper, we are concerned with handling instances admitting no complete composition solutions \emph{without} requiring  additional information, enhancements, or alterations of the domain of concern.
% %%

The rest of the paper is organized as follows.
In the next two sections, we introduce the composition framework as known in the literature. 
Besides providing the standard notion for exact compositions (complete solutions to the problem), we also introduce the notion of \emph{maximal compositions}, as controllers that can do as well as any other controller.
%%
% After that, we develop the main contribution of our work, namely, the notion of an \emph{optimal target approximation}, as the ``best'' alternative target behavior that can be fully realized in the system at hand. We demonstrate that using the best controller for an optimal approximate target in the original target amounts to using a maximal controller (for the original target), thus relating both notions and providing a correctness result for optimal approximations.
% 
After that, we develop the main contribution of our work, namely, the notion of \emph{optimal target approximations} as the best alternative target behaviors that can be fully realized in the system at hand. We demonstrate that ``importing'' controllers from optimal approximations amounts to using maximal controllers (for the original target), thus providing correctness for optimal approximations. 
In addition, we show that the imported controllers of an optimal approximation together realize the same set of traces as those realized by maximal controllers (together as well), thereby providing a completeness result.  
More importantly, we prove that optimal approximations are in fact unique (up to simulation equivalence), a very interesting and unexpected property. 
Finally, we describe how optimal approximate targets can be computed for the special case of deterministic systems (as, for example, in the context of service composition; e.g,~\cite{BerardiCDGP:IJFCS07,Balbianietal:SERVICES09}) by reducing the problem to ATL model checking, opening the door for advanced model checking tools.
We close the paper with a short discussion and conclusions.
An extended version of the paper, including proofs, can be found in the Appendix.

\section{The Behavior Composition Framework}\label{sec:behCompProblem}

% The behavior composition framework we are concerned with is that of 
% ~\cite {DeGiacomoS:IJCAI07,StroederPagnucco:IJCAI09,SardinaPDG:KR08}.
%%
In a behavior composition setting, a set of \emph{available behaviors} are meant to jointly bring about a \emph{virtual target behavior}~\cite {DeGiacomoS:IJCAI07,SardinaPDG:KR08,StroederPagnucco:IJCAI09}. 
%All behaviors are modelled using (non-deterministic) transition systems.
%%
%For simplicity, we do not discuss the so-called ``environment'', the shared space where behaviors are meant to execute. Still, all results to be presented can be easily adapted to deal with an environment.
%The behavior composition framework we are concerned with is similar to the one in \cite {DeGiacomoS:IJCAI07,StroederPagnucco:IJCAI09}. In a nutshell, a set of \emph{available behaviors} are meant to jointly bring about a virtual \emph{target behavior}. All behaviors are modelled using (non-deterministic) transition systems.
%%
We follow the composition framework in~\cite{SardinaPDG:KR08} with two minor modifications. For simplicity, we do not deal with the so-called environment, the shared space where behaviors are meant to execute. Nonetheless, all results presented here can be easily generalized to account for an environment. Second, we shall generalize target behaviors to non-deterministic transition systems.

% \vspace{-.3cm}
\paragraph{Behaviors}
A behavior stands for the operational model of a program or device.
In general, behaviors provide, step by step, the user a set of actions that it can perform (relative to its specification).
At each step, the behavior can be instructed to execute one of the legal actions, causing the behavior to transition to a successor state, and thereby providing a new set of applicable actions.

Formally, a \defterm{behavior} is a tuple $\B= \tup{B,\A,b_0,\varrho}$, where:%
\footnote{With no shared environment in this paper, behaviors are not equipped with guard conditions (as done in \cite{DeGiacomoS:IJCAI07,StroederPagnucco:IJCAI09}) and the set of actions $\A$ are included in their definitions. 
}
\begin{itemize}
\item  $B$ is the finite set of behavior's states;

\item $\A$ is a set of actions;  

\item $b_0 \in B$ is the initial state;
\item  $\varrho \subseteq B\times\A \times B$ is the behavior's transition relation, where $\tup{b,a,b'}\in\varrho$, or $b\goto{a}b'$ in $\B$, denotes that action $a$ executed in behavior state
$b$ may lead the behavior to successor state $b'$.
\end{itemize}

Note that we allow behaviors to be non-deterministic, that is, given a state and an action, the behavior may transition to more than one state.
This implies that one cannot know beforehand what actions will be available to execute after an action is performed, as the next set of applicable actions would depend on the successor state in which the behavior happens to be in.
Hence, we say that non-deterministic behaviors are only \emph{partially controllable}.
A \emph{deterministic} behavior is one where there is no state $b\in B$ and action $a \in A$ for which there exist two transitions $b\goto{a}b'$ and $b\goto{a}b''$ in $\B$ with $b' \not= b''$. A deterministic behavior is \emph{fully controllable}.
For the sake of legibility and easier notation, we shall assume, wlog, that behaviors capture non-terminating processes and hence do not have any terminating  state with no outgoing transition.\footnote{As customary, e.g., in LTL verification, this can be easily achieved by introducing ``fake'' loop transitions.}

% 

%%
% Notice that, given a state in a deterministic behavior and a legal action in that state, we always know exactly \emph{the} next behavior's state. 
% %%
% Thus, deterministic behaviors are \emph{fully controllable} through the selection of the next action to perform.
% %%
\vspace{-.3cm}
\paragraph{System and Enacted System} %A system stands for a collection of behaviors that are at disposal.
A system is a collection of behaviors at disposal.
Technically, an (available) \defterm{system} is a tuple $\S = \tup{\B_1,\ldots, \B_n}$, where $\B_i = \tup{B_i,\A_i,b_{i0}, \varrho_i}$, for $i \in \set{1,\ldots,n}$, is a behavior, called an \defterm{available behavior} in the system.

% The available behaviors in a given system are meant to act in an interleaved fashion. 
%%
To refer to the behavior that emerges from the joint execution of behaviors in a system, we use the notion of enacted system behavior. 
The \defterm{enacted system behavior} of an available system $\S$ (as above) is a tuple $\E_\S=\tup{S_{\S},\A, \set{1,\ldots,n},s_{\S0},\delta_\S}$, where:
\begin{itemize}
\item $S_\S =  B_1\times\cdots\times B_n$ is the finite set of $\E_\S$'s states;  when $s_\S=\tup{b_1,\ldots,b_n}$, we denote $b_i$ by $\beh_i(s_\S)$, for $i \in \set{1,\ldots,n}$;

\item $\A = \bigcup_{i=1}^n \A_i$ is the set of actions of $\E_\S$;

\item $s_{\S0}\in S_\S$ with $\beh_i(s_{\S0})=b_{i0}$, for $i \in \set{1,\ldots,n}$, is $\E_\S$'s initial state;

\item $\delta_\S\subseteq S_\S\times \A\times\set{1,\ldots,n}\times S_\S$ is $\E_\S$'s transition relation, where $\tup{s_\S,a,k,s_\S'} \in  \delta_\S$, or $s_\S \goto{a,k}  s'_\S$ in $\E_\S$, iff:
\begin{itemize}
\item  $\beh_k(s_\S)\goto{a}\beh_k(s_\S')$ in $\B_k$; and

\item $\beh_i(s_\S)=\beh_i(s_\S')$, for $i \in \{1,\ldots,n\} \setminus \{k\}$.
\end{itemize}
% \myi $\env(s_\S)\goto{a}\env(s_\S')$ in $\E$;
% \myii $\beh_k(s_\S)\goto{g,a}\beh_k(s_\S')$ in $\B_k$, with
%   $g(\env(s_\S))=\true$, for some $g \in G_k$; and
% \myiii $\beh_i(s_\S)=\beh_i(s_\S')$, for $i \in \{1,\ldots,n\} \setminus
% \{k\}$.
\end{itemize}

The enacted system behavior $\E_\S$ is technically the asynchronous product of the available behaviors.  
%%
%The presence of an index $k$ in transitions makes explicit which behavior is the one performing the action in the transition; all other behaviors remain still. 
The index $k$ in transitions makes explicit which behavior is performing the action in the transition---all other behaviors remain still.

% \vspace{-.3cm}
\paragraph{Target}
% - we define non-deterministic target here. We assume targets are always cyclic. Discuss and introduce $nop$ to generate cyclic targets from chains.
% Target requests transitions instead of actions.
%
A \defterm{target behavior} $\T = \tup{T,\A_T,t_0,\varrho_T}$ is a, possibly \textit{non-deterministic}, behavior that represents the desired functionality to be obtained (through the available system).
In contrast with all previous works, we allow for \emph{non-deterministic} target specifications. 
Nonetheless, the objective is \emph{not} to capture incomplete information, and hence partial controllability, of the target module, but to be able to accommodate  action requests carrying more ``information.'' This will come handy for our account of approximation.
Thus, in order to preserve the full controllability of the target, we shall consider requests in terms of  \emph{target transition}, rather than just actions.

% Nonetheless, the motivation for this more general account of target behaviors is not to accommodate incomplete knowledge of the target module 	(and hence allow for partially controllable targets), but to be able to specify more constrained modules. 
%%

\medskip
Informally, the behavior composition task is stated as follows: Given a system $\S$ and a target behavior $\T$, is it possible to (partially) control the available behaviors in $\S$ in a step-by-step manner---by instructing them on which action to execute next and observing, afterwards, the outcome in the behavior used---so as to ``realize" the desired target behavior. In other words, by adequately controlling the system, it appears as if one was actually executing the target module. (See next section for more details.)

As noted by \citea{DeGiacomoS:IJCAI07}, the behavior composition problem is related to planning (under incomplete information)~\cite{GhallabNT:04-Planning}, being both synthesis tasks, though here, we look for whom to delegate the next action at each step (whatever such action happens to be at runtime), rather than what those actions should be.

\begin{figure*}[!t]
\begin{center}
\resizebox{\textwidth}{!}{
\begin{tikzpicture}
 \tikzset{ActionStyle/.style = {font=\small}}
\begin{scope}[shift={(0cm,0cm)},node distance=2cm,auto]
\node[initial above,state]	(a0)                    {$a_0$};
\node[state]    			(a1) [right of=a0] 		{$a_1$};
\node[state]                (a2) [above right of=a1,shift={(0.35cm,-0.5cm)}]      {$a_2$};
\node[state]                (a3) [below right of=a1,shift={(0.35cm,0.5cm)}]      {$a_3$};

\path 	(a0) edge             	node[ActionStyle] 					{$\aMovie$}   (a1)
        (a1) edge[]          	node[ActionStyle,above,sloped]   
        							{$\aGame$}
								node[ActionStyle,below,sloped]   
        							{$\aWeb$}        							    (a2) 
        (a2) edge[bend right]   node[ActionStyle,sloped,above,near end] 					{$\aStop$}    (a0)
(a1) edge               node[ActionStyle,sloped,above]                  {$\aWeb$}   (a3)
(a3) edge [bend left=10]  node[ActionStyle]                  {$\aUnplug$}   (a0);

\node (name)[below of=a1,shift={(-.5cm,0.7cm)}]	
	{\labelfig{Game Device $\B_{G}$}};
\end{scope}

% AUDIO DEVICE
\begin{scope}[shift={(5cm,0cm)},node distance=1.7cm]

\node[initial above,state]		(b0)                    	{$b_0$};
\node[state]                    (b1) [right of=b0] 		    {$b_1$};
\node[state]                    (b2) [right of=b1]          {$b_2$};

\path 	(b0) 	edge    			node[ActionStyle,swap]	{$\aMusic$}		(b1)
        (b1) 	edge	            node[ActionStyle,swap]	{$\aRadio$}     (b2)
        (b2)    edge[bend right]    node[ActionStyle,swap]	{$\aStop$}      (b0);

\node (name)[below of=b1,shift={(-.5cm,0.8cm)}]		
			{\labelfig{Audio Device $\B_{A}$}};
\end{scope}

% LIGHT DEVICE
\begin{scope}[shift={(5cm,-2.7cm)},node distance=2.2cm]

\node[initial above,state]		(d0)                    	{$d_0$};
\node[state]                    (d1) [right of=d0] 		    {$d_1$};

\path 	(d0) 	edge[bend left]	   node[ActionStyle]	{$\aLightOn$}	(d1)
        (d1) 	edge   node[ActionStyle]	{$\aLightOff$}  (d0);

\node (name)[below of=d1,shift={(-1cm,1.4cm)}]		
	{\labelfig{Light Device $\B_{L}$}};
\end{scope}

% MOVIE DEVICE
\begin{scope}[shift={(0cm,-2.7cm)},node distance=1.8cm]

\node[initial above,state]      (c0)                        {$c_0$};
\node[state]                    (c1) [right of=c0]          {$c_1$};
\node[state]                    (c2) [right of=c1]          {$c_2$};

\path   (c0)    edge                node[ActionStyle,swap]  {$\aMovie$}     (c1)
        (c1)    edge                node[ActionStyle,swap]  {$\aRadio$}     (c2)
        (c2)    edge[bend right]    node[ActionStyle,swap]  {$\aStop$}      (c0);

\node (name)[below of=c1,shift={(-.5cm,1cm)}]       {\labelfig{Movie Device $\B_{M}$}};
\end{scope}

% TARGET
\begin{scope}[shift={(10cm,0cm)}, node distance=1.9cm]

\node[initial above,state]	(t0)                                     {$t_0$};
\node[state]                (t1) [right of=t0] 	                     {$t_1$};
\node[state]	            (t2) [right of=t1]	                     {$t_2$};
\node[state]	            (t3) [right of=t2]                       {$t_3$};
\node[state]	            (t4) [right of=t3]                       {$t_4$};

\path 	(t0) 	edge                node[ActionStyle,swap]  		{$\aLightOn$} 	        (t1)
        (t1) 	edge	            node[ActionStyle]	 	{\begin{tabular}{c}
                                                                $\aMovie$ \\ $\aMusic$ 
                                                             \end{tabular}   
                                                                } 	(t2)
        (t2) 	edge	            node[ActionStyle]		{\begin{tabular}{c}
                                                                $\aGame$ \\ $\aRadio$
                                                             \end{tabular}   } 
        				            node[ActionStyle,swap]		{\begin{tabular}{c}
                                                                $\aWeb$
                                                             \end{tabular}   } 
                                                            						(t3)
        (t3) 	edge				node[ActionStyle]   	{$\aStop$}              (t4)
        (t4) 	edge[bend right=30]	node[ActionStyle,swap]   	{$\aLightOff$}          (t0);

\node (name)[above of=t0,shift={(0.8cm,-0.7cm)}]			
		{\labelfig{Target  $\TEX$}};
\end{scope}

% TARGET APPROX
\begin{scope}[shift={(10cm,-2.5cm)}, node distance=2cm]

\node[initial above,state]  (t0)                                     {$u_0$};
\node[state]                (t1) [right of=t0]                       {$u_1$};
\node[state]                (t2) [above right of=t1,shift={(0.05cm,-0.35cm)}]                 {$u_2$};
\node[state]                (t3) [right of=t2]                       {$u_3$};
%\node[state]                (t4) [right of=t3]                       {$u_4$};
\node[state]                (t5) [below right of=t1,shift={(0.05cm,0.35cm)}]                 {$u_6$};
\node[state]                (t6) [right of=t5]                       {$u_7$};
%\node[state]                (t7) [right of=t6]                       {$u_7$};
\node[state]                (t8) [right of=t1]                       {$u_4$};
\node[state]                (t9) [right of=t8,xshift=0cm]          {$u_5$};
\node[state]                (t10) [right of=t9]                      {$u_{8}$};

\path   (t0)    edge                node[ActionStyle]       {$\aLightOn$}           (t1)
        (t1)    edge                node[ActionStyle]       {$\aMovie$}             (t2)
        (t2)    edge                node[ActionStyle]       {$\aGame$}              (t3)
        (t3)    edge[bend left=10]  node[ActionStyle,sloped,above]       {$\aStop$}              (t10)
        
        (t1)    edge                node[ActionStyle,swap]       {$\aMusic$}             (t5)
        (t5)    edge                node[ActionStyle]       {$\aRadio$}             (t6)
        (t6)    edge[bend right=5]  node[ActionStyle,sloped,above]       {$\aStop$}              (t10)

        (t1)    edge                node[ActionStyle]       {$\aMovie$}             (t8)
        (t8)    edge                node[ActionStyle]       {$\aRadio$}             (t9)
        (t9)    edge                node[ActionStyle]       {$\aStop$}             (t10)
        (t10)   edge[bend left=42]  node[ActionStyle,very near end]       {$\aLightOff$}          (t0)
        ;
\node (name)[above of=t0,shift={(1.3cm,-.7cm)}] 
	        {\labelfig{Target Approx $\aprxtgtEX$}};
\end{scope}

\end{tikzpicture} 
}
\end{center}
\caption{A smart house scenario with four available behaviors. Target $\TEX$ cannot be fully realized in the system, but its optimal approximation $\aprxtgtEX$ can.}
\label{fig:ambientIntel-system}
\end{figure*}
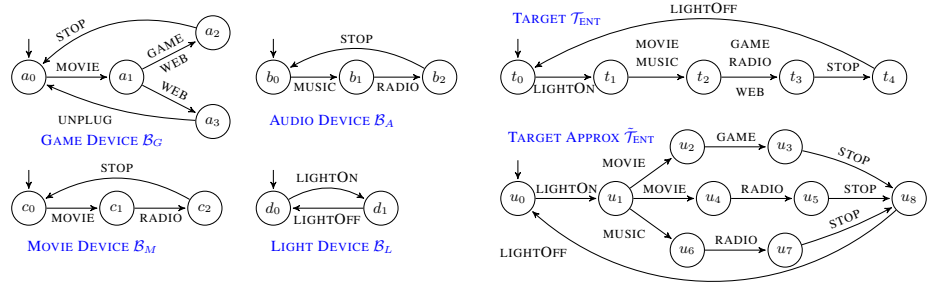

Figure~\ref{fig:ambientIntel-system} depicts a universal home entertainment system in a smart house scenario.
Target $\TEX$ encapsulates the desired functionality, which involves first
switching on the lights when entering the room, then providing various entertainment options (e.g., listening to music, watching movies, browsing the Web, etc.), and finally stopping active modules and switching off the lights.
There are four available devices installed in the house that can be used to bring about such desired behavior, namely, a game device $\B_G$, an audio device $\B_A$, a movie device $\B_M$, and the lights controller $\B_L$.
Note that action $\aWeb$ in the device $\B_G$ is non-deterministic, as it may bring the module into states $a_2$ or $a_3$. If the device happens to evolve to  state $a_3$, then, for some reason, it is not enough to stop the device to reset it: the device needs to be completely unplugged.

% Let us next make the composition problem definition precise.

% \begin{example}
% Figure~\ref{fig:ambientIntel-system} depicts a universal home entertainment system in a smart house scenario.
% %%
% Target $\TEX$ encapsulates the desired functionality, which involves first
% switching the lights on when entering the room, then providing various entertainment options (e.g., listening to music, watching movies, Web browsing the Web, etc.), and finally stopping active modules and switching off the lights.
% %%
% There are four available devices installed in the house that can be used to bring about such behavior, namely, a game device $\B_G$ capable of playing a movie, a game, or browsing the web, an audio device $\B_A$, a movie device $\B_M$, and a controller for the room's lights $\B_L$.
% %%
% Note that action $\aWeb$ in the device $\B_G$ is non-deterministic, as it may bring the module into state $a_2$ or $a_3$. If the device happens to evolve to the state $a_3$, then, for some reason, it is not enough to stop the device: it needs to be completely unplugged before functioning again.
% \end{example}

\section{Controllers and Compositions}\label{sec:controllers}

Next, we formally define what constitutes a solution for a behavior composition problem.
In doing so, we shall not only look at the problem from a binary perspective ---solvable vs unsolvable--but instead provide a \emph{qualitative} account of ``optimal'' solutions.
From now on, let $\S=\tup{\B_1,\ldots,\B_n}$ be an available system and $\T= \tup{T, \A, t_0, \varrho_T}$ be a target behavior to be realized on $\S$.

% \vspace{-.3cm}
\paragraph{Controller} 
A controller is a component able to activate, stop, and resume any of the available behaviors, and to instruct them to execute an (allowed) action.
The controller has \textit{full observability} on the available behaviors; that is, it can keep track (at runtime) of their current states---if details have to be hidden, this can be done by means of non-determinism within the abstract behaviors exposed.
%%
% As argued by~\citea{DeGiacomoS:IJCAI07}, full observability is the natural choice in this context. Since available behaviors are already suitable abstractions for \textit{actual} modules, if details have to be hidden, this can be done by means of non-determinism within the abstract behaviors exposed.
%%
%
%Roughly speaking, we look for a controller that can ``realize" (i.e., implement) the target behavior by suitably operating the available behavior modules. 
%Informally, the task is to look for a controller than can ``realize" (i.e., implement) the target behavior by suitably operating the available behavior modules. 
%
%%
% Unlike previous work, except for \citea{YadavSardina:AAMAS11}, we shall look for controllers that can achieve such a realization \emph{to the maximum degree possible}; thus providing a solution account for problems that do not admit complete solutions.

To formally define controllers and solutions, we rely on the notions of traces and histories. 
A \defterm{trace} for a given enacted system $\E_\S=\tup{S_{\S},\A, \set{1,\ldots,n},s_{\S0},\delta_\S}$ is a, possibly infinite, sequence of the form $s^0 \goto{a^1,k^1} s^1 \goto{a^2,k^2}\cdots$ such that \myi $s^0=s_{\S0}$; and \myii $s^{j} \goto{a^{j+1},k^{j+1}} s^{j+1}$ in $\E_\S	$, for all $j > 0$.
A  \defterm{history} is just a finite prefix $h = s^0 \goto{a^1,k^1} \cdots \goto{a^\ell,k^\ell}s^\ell$ of a trace. We denote $s^\ell$ by $\last(h)$, $\ell$ by $\card{h}$ (i.e., the length of $h$), and sequence $a^1 \cdot \ldots \cdot a^\ell$ as $\projact{h}$ (i.e., the projection on actions).
Traces and histories can also be defined for a behavior $\B$ in a similar fashion: behavior traces have the form $s^0 \goto{a^1} s^1 \goto{a^2}\cdots$ such that \myi $s^0=b_0$; and \myii $s^{j} \goto{a^{j+1}} s^{j+1}$ in $\B$, for all $j > 0$.
We use $\H_{\S}$ and $\H_\B$ to denote the set of system histories (i.e., histories of $\E_\S$) and histories of behavior $\B$, respectively.

A \defterm{controller} for target $\T$ on system $\S$ is a partial function $C:\H_\S \times (T \times \A \times T)  \mapsto \set{1,\ldots,n}$,
% \[
% 	C:\H_\S \times (T \times \A \times T)  \mapsto \set{1,\ldots,n},
% \]
%%
which, given a system history $h\in\H_\S$ and a requested target transition $\tup{t,a,t'} \in \varrho_T$, returns the index of an available behavior to which the action $a$ is delegated for execution.
%
%selects an available behavior---actually, returns its index---to delegate action $a$ to for execution.
%%
% \footnote{For technical convenience, a special value $u$ (``undefined'') may be returned, thus making $P$ a total function which returns a value even for irrelevant histories or non-feasible requests.}
% %%
%
For legibility, we shall write $C(h,t_1\goto{a}t_2)$ to compactly denote $C(h,t_1,a,t_2)$.
Note here the slight departure form previous notions of controllers (e.g., \cite{DeGiacomoS:IJCAI07,StroederPagnucco:IJCAI09,SardinaPDG:KR08}), in that a controller now receives a complete target transition as the next request, not just an action.
While this has no impact when dealing with deterministic targets, it guarantees full controllability for nondeterministic ones.

Intuitively, a controller (fully) realizes a target behavior if for every trace (i.e., run) of the target, at every step, the controller returns the index of an available behavior that can perform the requested action.
%%
% As noted by \citea{DeGiacomoS:IJCAI07}, the behavior composition problem is related to planning~\cite{GhallabNT:04-Planning}, being both synthesis tasks, though here, we look for whom to delegate the next action at each step (whatever such action happens to be at runtime), rather than what those actions should be.
%%
Formally, one first defines when a controller $C$ \defterm{realizes a trace} of the target $\T$. Though not required for this paper, the reader is referred to~\cite{DeGiacomoS:IJCAI07,SardinaPDG:KR08} for details on how to formally characterize trace realization.
We denote $\Delta^C_{\tup{\S,\T}}$ the set of traces of $\T$ that controller $C$ is able to realize in system $\S$.
Then, a controller $C$ realizes the target behavior $\T$ \emph{iff} it realizes all its traces. In that case, $C$ is said to be an \defterm{exact composition} for target $\T$ on system $\S$.

% Techniques for checking the existence and synthesis of a exact compositions are known \cite{DeGiacomoS:IJCAI07,AAAI07,SardinaPDG:KR08,StroederPagnucco:IJCAI09}. 
% %%
% However, as far as we know, except for \cite{YadavSardina:AAMAS11}, which we will discuss later on, no existing approach is able to handle problem instances which do not have exact compositions. The point is that the notion of exact composition is not enough to either capture ``degrees'' of realization or to compare different non-complete controllers.	

% 
% In fact, there is so far no notion of ``optimal'' realization that tell us what the ``best'' possible realization of a target system in a given system is. 

\newcommand{\domin}{\geq}
\newcommand{\Max}{\operatorname{\emph{MAX}}}
\renewcommand{\Max}{\operatorname{\textsc{MaxComp}}}

\medskip
Now, suppose we are given a target behavior $\T$ and an available system $\S$, and that, as expected in many domains, there is no exact composition for $\T$ on $\S$---the target cannot be \emph{completely} realized in the system. 
This is indeed the case in our example, as there is no exact composition for $\TEX$ in the house system.
Merely returning a negative ``no solution'' outcome is highly unsatisfactory. 
The question then is: what does it mean for a controller $C_1$ to achieve ``a better realization'' of $\T$ on $\S$ than controller $C_2$?

% To answer such a question in a qualitative manner and without enriching the framework, we rely on the extent at which controllers are able to honour arbitrary long set of target requests.
%%
To answer such a question in a qualitative manner, we rely on the extent at which controllers are able to honour arbitrary long set of target requests.
We say that controller $C_1$ \defterm{dominates} controller $C_2$, denoted $C_1 \domin C_2$, \emph{iff} $\Delta^{C_2}_{\tup{\S,\T}} \subseteq \Delta^{C_1}_{\tup{\S,\T}}$---$C_1$ can honour all  request sequences that $C_2$ can honour, and possibly more.
As usual, $C_1>C_2$ is equivalent to $C_1 \geq C_2$ but $C_2 \not\geq C_1$, that is, 
$\Delta^{C_2}_{\tup{\S,\T}} \subset \Delta^{C_1}_{\tup{\S,\T}}$.
A controller $C$ is said to be a \defterm{maximal composition} (for a target on a system) \emph{iff} for every other controller $C'$, if $C' \domin C$, then $C \domin C'$ (or equivalently $C' \not> C$). In other words, maximal compositions are those for which there is no other controller that can realize strictly more runs of the target behavior in the system.
We use $\Max(\S,\T)$ to denote the set of all maximal compositions for target $\T$ on system $\S$.

Consider the following two controllers for our smart house.
Whereas controller $C_1$ allocates all requests to the light device $\B_L$, controller $C_2$ delegates media and light requests to the audio $\B_A$ and light  $\B_L$ devices, respectively.
Then, $C_1$ realizes just one target trace, that is, $\Delta^{C_1}_{\tup{\S,\T}} =\set{t_0\goto{\aLightOn}t_1}$.
On the other hand, $C_2$ realizes such a trace as well as trace $t_0\goto{\aLightOn}t_1\goto{\aMovie}t_2 \goto{\aRadio} t_3 \goto{\aStop} t_4$ (and all its prefixes).
Therefore, $\Delta^{C_1}_{\tup{\S,\T}} \subset \Delta^{C_2}_{\tup{\S,\T}}$ and  $C_2>C_1$ holds.
The reader may notice that even better controllers than $C_2$ exist when all four behaviors are used.

% \begin{example}
% Consider two controllers $C_1$ and $C_2$ for the smart house scenario.
% %%
% Whereas controller $C_1$ allocates all transition requests to the light device $\B_L$, controllers $C_2$ delegates media requests to the audio device $\B_A$ and lights requests to the light device $\B_L$.
% %%
% Then, $C_1$ realizes just one trace, that is, $\Delta^{C_1}_{\tup{\S,\T}} =\set{t_0\goto{\aLightOn}t_1}$.
% %%
% On the other hand, $C_2$ realizes such trace as well as trace $t_0\goto{\aLightOn}t_1\goto{\aMovie}t_2 \goto{\aRadio} t_3 \goto{\aStop} t_4$ (and all its prefixes).
% %%
% Therefore, $\Delta^{C_1}_{\tup{\S,\T}} \subset \Delta^{C_2}_{\tup{\S,\T}}$ and  $C_2>C_1$ holds.
% %%
% Clearly, even better controllers exist when all four behaviors are used.
% \end{example}

As expected, whenever a behavior composition problem admits an exact composition---the target is fully realizable---the set of exact compositions coincides with that of maximal compositions.
When full realizations are impossible, though, maximal compositions capture the best controllers that one could hope for. 
%%
% In the next section, we shall look at ``approximate'' solutions from a different perspective that is arguably more intuitive and computationally more amenable, than dealing with controller functions.

% Although we now have the new notion of maximal compositions (based on the standard concept of dominance) to deal with cases where  is not possible, it is not clear

\section{Target Approximation}\label{sec:targetApproximation}

Whereas maximal compositions, as defined above, provide a way of handling  instances with no exact solution, they do not convey useful insights on how well such instances can be solved. Even if we are given the set of traces that a maximal composition realizes, it will be difficult to reconstruct what it means in terms of the problem specification. As a consequence, using a maximal non-exact composition may yield dead-end executions where no further actions can be honoured.
What is more, while there are various techniques to construct exact compositions (e.g.,~\cite{DeGiacomoS:IJCAI07,SardinaDG:ICAPS08,StroederPagnucco:IJCAI09}), it is far from clear how to build maximal composition controllers.

So, in this section, we will look at ``approximation'' from a different perspective that is arguably more intuitive and computationally more amenable than dealing with controller functions, namely, we are concerned with what parts of the target can in fact be brought about. 	
%%
% More concretely, given a behavior composition problem $\tup{\S,\T}$---that is, realize $\T$ on $\S$---we shall look for an (approximate) target behavior $\aprxtgt$, and a controller $\aprxcntrl$, that can indeed be fully realized on $\S$ (by controller $\aprxcntrl$) and such that $\aprxtgt$  is ``reasonably'' close to the original module $\T$. 
%%
More concretely, we are interested in the following task:
\begin{quote}
\emph{Given an available system $\S$ and a target behavior $\T$, find an (approximate) target behavior $\aprxtgt$  that can be fully realized on $\S$ (by some controller $\aprxcntrl$) and such that $\aprxtgt$  is ``as close as possible'' to the original target behavior $\T$. }
\end{quote}

We call this the \emph{approximate behavior composition problem}.
Once an approximate target $\aprxtgt$ is obtained, one may either use such new target directly or consider ``importing'' its exact compositions into the original target module $\T$.
Hopefully, in the latter case, the imported controllers will turn out to be the best possible controllers for the original target.
These are arguably the main ideas of our work and what we shall develop below.
Before doing so, we should point out that defining approximate targets based merely on trace/language inclusion is not sufficient. While two targets may yield exactly the same sequences of requests, one may accept an exact composition while the other may not.  In our smart house scenario, for instance, the two sequences $\aLightOn \cdot \aMovie \cdot \aGame \cdot \aStop$ and $\aLightOn \cdot \aMovie \cdot \aRadio \cdot \aStop$ may be realized by the same controller for the approximation $\aprxtgtEX$, but not for the original target $\TEX$.

In order to capture approximate targets, we make use  of the formal notion of \emph{simulation}~\cite{Miln71}.
A simulation relation captures the similarity in the behavior of two transition systems.
Intuitively, a (transition) system $S_1$ ``simulates'' another system $S_2$ if $S_1$  is able to \textit{match} all of $S_2$'s moves.
We make this precise for our (target) behaviors as follows.
Let $\T_i=\tup{T_i, \A, t_{i0}, \varrho_i}$, where $i\in \set{1,2}$, be two target behaviors. 
A \defterm{simulation relation} of $\T_2$ by $\T_1$ is a relation $\Sim \subseteq T_2 \times T_1$ such that $\tup{t_2,t_1} \in \Sim$ implies that for every transition $\tup{t_2,a,t_2'} \in \varrho_2$ in $\T_2$, there exists a transition $\tup{t_1,a,t_1'} \in \varrho_1$ in $\T_1$ such that $\tup{t_2',t_1'} \in  \Sim$.
We say that a state $t_2 \in T_2$ is \defterm{simulated} by a state $t_1 \in T_1$ (or $t_1$ simulates $t_2$), denoted $t_2 \preceq t_1$, \emph{iff} there exists a simulation relation $\Sim$ of $T_2$ by $T_1$ such that $\tup{t_2,t_1} \in \Sim$. Observe that relation $\preceq$ is itself a simulation relation (of $\T_2$ by $\T_1$), and in fact, it is the largest simulation relation, in that all simulation relations are contained in it. 
Informally, $t_2 \preceq t_1$ means that $t_1$ in $\T_1$ can ``mimic'' all moves of $t_2$ in $\T_2$, and that this property is propagated in their corresponding successor states.
We say that a target behavior $\T_1$ \defterm{simulates} target behavior $\T_2$, denoted $\T_2 \preceq \T_1$, if it is the case that $t_{20} \preceq t_{10}$, that is, their initial states are in simulation and, as a result, $\T_1$ can always mimic $\T_2$ from the start.
In our example, $t_2$ and $t_1$ in $\TEX$ simulate states $u_4$ and $u_1$, respectively, in $\aprxtgtEX$ (i.e., $u_4 \preceq t_2$ and $u_1 \preceq t_1$), but not the other way around (i.e., $t_2 \not \preceq u_4$ and $t_1 \not \preceq u_1$). 
Two targets are said to be \defterm{simulation equivalent}, denoted $\T_1 \sim \T_2$, whenever they simulate each other.

We then argue that a qualitative comparison of target approximations can be achieved based on their simulation ``hierarchy'' (see that $\preceq$ is a pre-order).
We say that a target behavior $\aprxtgt$ \defterm{approximates} target $\T$ on system $\S$ (or $\aprxtgt$ is an approximation of $\T$ on $\S$) \emph{iff} $\aprxtgt \preceq \T$ and there is an exact composition for $\aprxtgt$ on $\S$ (i.e., $\aprxtgt$ is simulated by $\T$ and it can be fully realized on available system $\S$).

Despite being fully solvable, an approximation will generally provide ``\emph{less}'' than the original target. 
First, an approximation may be missing certain executions altogether. In the smart house scenario, approximation $\aprxtgtEX$ does not account for the action sequence $\aLightOn \cdot \aMusic \cdot \aGame \cdot \aStop \cdot \aLightOff$.
Second, an approximation may require the user to commit earlier to future possible request choices. In that sense, a user of target $\aprxtgtEX$ needs to decide   when requesting $\aMovie$ in state $u_1$ if she will later play a $\aGame$ or listen to $\aRadio$. Notice such extra ``temporal'' information is not required at state $t_1$ in original target $\TEX$.
It is exactly to accommodate this feature that we have departed from the standard view of deterministic targets.

Of course, between full realization and the trivial empty approximation, there lies a whole spectrum of approximating targets.
Among these, we are interested in those that are ``closest'' to the original target, in that the minimum possible is given up.
We say that a target behavior $\aprxtgt$ is an \defterm{optimal approximate} of target $\T$ on system $\S$ \emph{iff}:
\begin{enumerate}
  \item  $\aprxtgt$ is an approximation of $\T$ on $\S$; and 
  \item there is no target behavior $\aprxtgt'$ that approximates $\T$ on $\S$ such that $\aprxtgt \prec \aprxtgt'$, that is, $\T$ cannot be approximated by a strictly more general target module.
\end{enumerate}

Intuitively, an optimal target approximation is a maximal representation of those aspects of the original target that can be completely implemented.
When the target behavior does admit a full realization in the system, the optimal approximation is then expected to represent the target module in all its extent. 
%%
% This is what the following result states.

\begin{theorem}\label{theo:optaprox_solvable}
Suppose there is an exact composition for target $\T$ on system $\S$.
% Let $\T$ be a target behavior and $\S$ be an available system, such that there exists an exact composition for $\T$ on $\S$. 
% %%
Then, $\aprxtgt$ is an optimal approximation of $\T$ on $\S$ iff $\aprxtgt \sim \T$.
\end{theorem}

% Importantly, %also, 
% there can never be more than one way of optimally approximating a given target. 
Importantly, there can only be one way of optimally approximating a given target.

\begin{theorem}\label{theo:optaprox_unique}
An optimal approximation $\aprxtgt$ of a target $\T$ on a system $\S$ is unique upto simulation equivalence.
\end{theorem}

We observe that, for non-deterministic transition systems, simulation is a stronger measure of equivalence than language inclusion~\cite{GirardPappas:IEEE2007}. Therefore, if a target $\aprxtgt$ approximates another target $\T$,  then the action request sequences resulting from the traces of $\aprxtgt$ will be a  subset of those produced by $\T$.
It follows then that if $\aprxcntrl$ is an exact composition for $\aprxtgt$, then $\aprxcntrl$ ought to be able to handle a subset of $\T$'s request sequences.

\subsection{Imported Controllers}\label{secsub:induced_controllers}

In contrast with maximal controllers, optimal approximations are specified in the \emph{same} language as the original problem.
%%
%The user can thus decide to request actions as per the new (approximate) target and be guaranteed her requests will always be honored.
The user can thus decide to request actions as per the new (approximate) target with guaranteed full realizability.
Nonetheless, one may still ask in which sense these solutions are ``correct.''  
To answer that, we show that using an exact composition for an optimal approximation amounts to using a maximal composition for the original target.
To that end, we define what it means to ``import" a controller $C_{\T'}$ designed for one target module $\T'$ into another target module $\T$.

We start by defining the family of functions that are meant to explain sequences of action requests in a target.
Informally, the function $\expl_\T(\sigma)$ outputs a history of the target $\T$ compatible with the given sequence of actions $\sigma$.
Formally, a function $\expl_\T : \A^* \mapsto \H_\T$ is a \defterm{target explanatory} function for a target $\T$ if for any action sequence $\sigma = a^1 \cdot \ldots \cdot a^\ell \in \A^*$, with $\ell \geq 0$,  it is the case that $\expl_\T(\sigma) = t^0 \goto{a^1} \cdots \goto{a^\ell} t^\ell \in \H_\T$.
%%
%Note that, 
In general, there will be many of such functions, since the same sequence of action requests can arise from different runs of a non-deterministic target.
For instance, sequence $\aLightOn \cdot \aMovie$ can be explained in two ways on target $\aprxtgtEX$, namely, via histories 
$u_0\goto{\aLightOn} u_1 \goto{\aMovie} u_2$ and $u_0\goto{\aLightOn} u_1 \goto{\aMovie} u_4$.

% In addition we define the so-called \emph{target mapping} functions, functions meant to translate sequences of requests from one target behavior 	to equivalent requests from another, often more general, target.
% %%
% As expected, we rely on simulation to capture such equivalence.
% %%
% A \defterm{target mapping} function from behavior $\T_1$ to behavior $\T_2$ if  a function $\map : \H_{\T_1} \mapsto \H_{\T_2}$ such that for any history $\tau = t_1^{0} \goto{a^1} \cdots \goto{a^\ell} t_1^\ell$ of $\T_1$, with $\ell \geq 0$, $\map(\tau)$ returns a history $t_2^{0} \goto{a^1} \cdots \goto{a^\ell} t_2^{\ell}$ of $\T_2$, if any, such that $t_1^i \preceq t_2^i$, for every $i \in \set{0,\ldots,\ell}$, and undefined otherwise.
% %%
% Roughly speaking $\map(\tau)$ represents the same kind of request sequence that $\tau$ but in target module $\T_2$ rather than in $\T_1$.
% %%
% Observe that due to nondeterminism, a history of $\T_1$ could map to more than history of $\T_2$, so a family of target mapping functions may exist.
% %%
% \textbf{For example, both the traces $u_0\goto{\aLightOn} u_1 \goto{\aMovie} u_2$ and $u_0\goto{\aLightOn} u_1 \goto{\aMovie} u_8$ in $\T_{R_2}$ map to a single trace $t_0\goto{\aLightOn} t_1 \goto{\aMovie} t_2$ in $\T_{R_1}$. NOTE: WE SAY IT MAPS TO MANY, AND HERE WE SAY TO A SINGLE? FLOW HAS TO BE CONSISTENT, NO GUESSING... ;-)
% }

Using target explanatory functions, we next characterize the set of so-called \emph{induced} controllers. Suppose we have a controller $C_{\T'}$ for a target $\T'$ (on a system $\S$). 
An induced controller (from controller $\C_{\T'}$) for a target behavior $\T$ is one that handles requests from $\T$ as if they were requests issued as per module $\T'$. 
Recall that a controller for a system $\S$ outputs the behavior index to which a given transition-action request is delegated to at a certain system history.
Formally, then, we say that $C^{\T'}_\T$ is an  \defterm{induced controller} (from controller $C_{\T'}$ on target  $\T'$) for target $\T$ over system $\S$ if there exists a target explanatory function $\expl_{\T'}(\cdot)$ for $\T'$ such that for every system history $h \in \H_{\S}$ and transition $t_1 \goto{a} t_2$ in $\T$, the following holds (recall that $\projact{h}$ denotes the sequence of actions in history $h$):  
%%
% \[
% C^{\T'}_{\T}(h,t_1 \goto{a} t_2) =
%     \C_{\T'}(h,t_1'\goto{a}t_2'), 
%     \text{ where $  
%         \expl_{\T'}(\projact{h} \cdot a)= 
%         t^0 \goto{a^1} \cdots \goto{a^{\card{h}}} t_1' \goto{a} t_2'
%         $}.
% \]
\[
C^{\T'}_{\T}(h,t_1 \goto{a} t_2) \! = \! 
\begin{cases}
          \C_{\T'}(h,t_1'\goto{a}t_2') & 
          \expl_{\T'}(\projact{h} \cdot a) \!=\! 
        t^0 \goto{a^1} \!\cdots\!\goto{a^{\card{h}}} t_1' \goto{a} t_2'     \\
           \text{undefined} & 
           \text{$\expl_{\T'}(\projact{h} \cdot a)$ is undefined} \\
\end{cases}
\]         
That is, $\T$'s request $t_1 \goto{a} t_2$ is delegated at history $h$ as controller $C_{\T'}$ would delegate request $t_1' \goto{a} t_2'$ from target $\T'$ if $h$'s requests leave target $\T'$ in state $t_1'$ and the current requested action $a$ is indeed explained by transition request $t_1' \goto{a} t_2'$ in $\T'$.   
When there is no explanation in the $\T'$---$\expl(\cdot)$ is undefined---the induced controller is left undefined.
Note that different ways of explaining original target's sequences of requests (i.e., different explanatory functions) yield different induced controllers.

Finally, an \emph{imported} controller is a maximal (i.e., non-strictly dominated) controller within the family of induced controllers---the ``best''  induced controllers.
Technically, the set of \defterm{imported controllers} from $C$ on $\T$ into target $\T'$, denoted $\Omega^{\T'}_{\tup{C,\T} }$ is the set of all controllers $\hat{C}$ for $\T'$ such that \myi $\hat{C}$ is an induced controller from $C$ on target $\T$ for $\T'$; and \myii there is no other induced controller $C'$ such that $C' > \hat{C}$.
%%

%  (i.e., $\Delta_{\tup{\S,\T}}^{C} \subset \Delta_{\tup{\S,\T}}^{C'}$; $C'$ can realize more traces than $C$).
%%	
% We say that a set of controllers $\Omega^1$ is said to be \emph{as good as} another set of controllers $\Omega^2$ if there is no controller in $\Omega^1$ that is dominated by some controller in $\Omega^2$, that is, there is no $C_1 \in \Omega^1$ and $C_2 \in \Omega^2$ such that $C_2 > C_1$ (i.e., $\Delta_{\tup{\S,\T}}^{C_1} \subset \Delta_{\tup{\S,\T}}^{C_2}$).

First, we show that better target approximations amount to better, or more precisely ``never worse,'' imported controllers. 

\begin{theorem}\label{th:optimality}
% Let $\S$ be an available system and $\T$ a target behavior.
%%
Let ${\aprxtgt_1}$ and $\aprxtgt_2$ be two target approximations of target $\T$ on system $\S$, and let $\tilde{C_{1}}$ and $\tilde{C_{2}}$ be exact compositions of $\aprxtgt_1$ and $\aprxtgt_2$, resp.
Suppose also that ${\aprxtgt_2} \preceq {\aprxtgt_1}$ (i.e, ${\aprxtgt_1}$ simulates ${\aprxtgt_2}$).
Then, for every controller $C_1 \in \Omega^{\T}_{\tup{\tilde{C_{1}},\aprxtgt_1}}$, there is no controller $C_2 \in \Omega^{\T}_{\tup{\tilde{C_{2}},\aprxtgt_2}}$ such that $C_2 > C_1$ holds. 

% 
% \begin{Proofsk}
% Assume controllers $C_1$ and $C_2$ as above such that $C_2 > C_1$.
% %%
% Let $\expl_1$ and $\expl_2$ be the target explanatory functions that $C_1$ and $C_2$ are built upon, resp.
% %%
% Now, consider a target explanatory function $\expl_1'$ for $\aprxtgt_1$ such that $\expl_1'(h)$ simulates $\expl_2(h)$ state-wise (i.e., at each step). 
% %%
% % That is, for all histories $h \in \H_{\S}$ if $\expl_2(h) = t_2^0\goto{a^1}\cdots t_2^{|h|}$ then, $\expl_1'(h) = t_1^0\goto{a^1}\cdots t_1^{|h|}$ such that for all $i\leq |h|$ it is the case that $t_2^i \preceq t_1^i$.
% %%
% Note such function $\expl_1'$ exists since $\aprxtgt_1$ simulates $\aprxtgt_2$.
% %%
% Next, consider the imported controller $C_1' \in \Omega^{C_{\aprxtgt_1}}_{{\aprxtgt_1} \to \T}$ built upon target explanatory function $\expl_1'$. It is not hard to prove that, because traces obtained using $\expl_1'$ simulate those obtained using $\expl_2$, $C_1' \geq C_2$ holds (i.e., $C_1'$ dominates $C_2$).
% %%
% Since, by assumption, $C_2 > C_1$, it follows that $C_1' > C_1$, an contradiction since $C_1$ is not strictly dominated by any induced controller from $\aprxcntrl$ to $\T$.
% 
% \end{Proofsk}
\end{theorem}

In other words, if $\aprxtgt_1$ is as good an approximation as $\aprxtgt_2$, then $\aprxtgt_1$'s imported controllers will not be worse than those imported from $\aprxtgt_2$. 
More importantly, the next result demonstrates that importing  controllers from an \emph{optimal} approximation yields maximal compositions (for the original target being approximated), and that, together, they account for every trace of the original target that could ever be realized. In other words, $\Omega^{\T}_{\tup{\tilde{C},\aprxtgt}}$ is sound and ``complete.''

\begin{theorem}\label{theo:imported_are_max}
% Let $\S$ be an available system and $\T$ a target behavior. 
%%
Let $\aprxtgt$ be an optimal approximation of target $\T$ on system $\S$, and $\tilde{C}$ be an exact composition for $\aprxtgt$.
Then, 
\begin{itemize}
  \item For all $C \in \Omega^{\T}_{\tup{\tilde{C},\aprxtgt}}$, it holds that ${C} \in \Max(\S,\T)$; and
  
  \item $\bigcup_{C \in \Omega^{\T}_{\tup{\tilde{C},\aprxtgt}}} \Delta_{\tup{\S,\T}}^C = \bigcup_{C \in \Max(\S,\T)} \Delta_{\tup{\S,\T}}^C$, that is, all imported controllers account together for all realizable target traces.
\end{itemize}
\end{theorem}

These two results are important in that they establish the relationship between approximating the target and optimizing its controller:
optimizing targets implies optimizing controllers.
%
% An interesting observation is that, in the case of \emph{compact} optimal approximations, Theorem~\ref{theo:imported_are_max} can be further extended to state that the set of imported controllers is also \emph{non-redundant}, in that two imported controllers may not yield the same set of realized traces. 
% %%
% Roughly speaking, a compact behavior is one that has no redundancy: if there are two transitions $t_1 \goto{a} t_2$ and $t_1 \goto{a} t_2'$, there must then be some trace from $t_2$ that cannot be produced from $t_2'$, and vice-versa.
% %%
% Compactness of a behavior can be formally captured by appealing to bisimulation with itself.
%
%
%
%\footnote{We omit this result here, but will be added to the final version.} 
%
A direct and expected consequence of Theorems~\ref{theo:optaprox_solvable} and ~\ref{theo:imported_are_max} is that if the optimal approximation is simulation equivalent to the target, then every imported controller from such approximation is in fact an exact composition.

% A direct, and expected, consequence is that if an optimal approximation bisimulates the target module, then there is an exact composition for the composition problem at hand.
% 
% \begin{corollary}
% Let $\aprxtgt$ be an optimal target approximation of target $\T$ on system $\S$ such that $\aprxtgt \cong \T$, and let $\aprxcntrl$ be an exact composition for $\aprxtgt$ on $\S$. 
% %%
% Then, every imported controller in $\Omega^{C_{\aprxtgt}}_{\aprxtgt \to \T}$ is an exact compositions for $\T$ on $\S$.
% % 
% % \begin{Proofsk}
% % Since $\aprxtgt $ has an exact composition and $\aprxtgt$ can mimic $\T$ (due to $\T \preceq \aprxtgt$), $\T$ has an exact composition too: just delegate requests as done with $\aprxtgt$. 
% % %
% % Let $C$ be such exact composition for $\T$ in $\S$, and assume next that there exists $C' \in \Omega^{C_{\aprxtgt}}_{\aprxtgt \to \T}$ such that $C'$ is \emph{not} an exact composition for $\T$ in $\S$.
% % %%
% % Hence, $C > C'$ follows, a contradiction since $C' \in \Max(\S,\T)$ due to Theorem~\ref{theo:imported_are_max}.
% % \end{Proofsk}
% \end{corollary}

% \input{5-ImplSafetyGames}
%%%%%%%%%%%%%%%%%%%%%%%%%%%%%%%%%%%%%%%%%%%%%%%%%%%%%%%%%%%%%%%%%%%%%
% \section{Computing Optimal Target Approximations for Deterministic Available Systems}\label{sec:solution}
\section{Computing Optimal Approximations for Deterministic Systems}\label{sec:solution}
%%%%%%%%%%%%%%%%%%%%%%%%%%%%%%%%%%%%%%%%%%%%%%%%%%%%%%%%%%%%%%%%%%%%%

\newcommand{\coalition}[1]{\ensuremath{\langle\!\langle #1 \rangle\!\rangle}}
\newcommand{\coalitione}[2]{\coalition{#1}_{#2}}

\newcommand{\extract}{\propername{ExBeh}}

\newcommand{\ltl}{\textsc{ltl}\xspace}
\newcommand{\tlv}{\textsc{tlv}\xspace}
\newcommand{\anzus}{\textsc{Anzu}\xspace}
\newcommand{\mocha}{\textsc{Mocha}\xspace}
\newcommand{\lily}{\textsc{Lily}\xspace}
\newcommand{\mcmas}{\textsc{mcmas}\xspace}
\newcommand{\nugat}{\textsc{nugat}\xspace}

\newcommand{\adam}{system\xspace}
\newcommand{\eve}{controller\xspace}

\newcommand{\ind}{\name{k}\xspace}
\newcommand{\fail}{\name{fail}\xspace}

\newcommand{\dummy}{\ensuremath{\sharp}}
\newcommand{\Post}{\propername{Post}}
\newcommand{\bel}{\ensuremath{\mathbb{B}_{\S}}}

\newcommand{\contr}{\name{\textsf{contr}}}
\newcommand{\targetagt}{\name{\textsf{\small tgt}}}
\newcommand{\state}{\name{state}}
\newcommand{\error}{\name{error}}

%\newcommand{\Poss}{\propername{Poss}}
% In this section, we adopt a more pragmatic perspective and focus on finding effective ways for dealing with non-solvable composition instances.
%%

Various techniques have been used to actually solve classical behavior composition problems, including PDL satisfiability~\cite{DeGiacomoS:IJCAI07}, direct search-based approaches~\cite{StroederPagnucco:IJCAI09}, LTL/ATL synthesis~\cite{SardinaDG:ICAPS08,DeGiacomoFelli:AAMAS10}, and computation of special kind of simulation relations~\cite{SardinaPDG:KR08,BerardiCDGP:IJFCS07}.
Unfortunately, all those techniques synthesize \emph{exact} composition controllers. In the context of our work, we are interested in \emph{computing optimal target approximations} instead.
We show how this can be effectively done for the special case of \emph{deterministic} available behaviors, as in the case of service composition~\cite{BerardiCDGP:IJFCS07,Balbianietal:SERVICES09}.

% In this section, we explain how this can be done by resorting to model checking techniques applied to so-called \emph{safety-games}~\cite{Asarin-etal:HS95,Pitterman-etal:JCSS11-GR1Synthesis}. 
% %%
% The motivation behind  is the availability of software tools, such as \tlv~\cite{PnueliShahar:CAV96-TLV}, \nugat,\footnote{\url{https://es.fbk.eu/index.php?n=Tools.NuGaT}} \anzus~\cite{Jobstmann-etal:CAV07-ANZU}, and , providing effective procedures for strategy computation and convenient languages for representing the problem instance in a modular and high-level manner. 
% %%

\citea{DeGiacomoFelli:AAMAS10} has shown that the controller generator (i.e., a structure representing all exact compositions) can be synthesised  by resorting to Alternating-time Temporal Logic (ATL) model checking.
ATL~\cite{AluraHK:JACM02-ATL} is a logic for reasoning about the ability of group of agents (i.e., coalitions) in multi-agent game structures. 
%%%
The advantages of reducing the composition problem to that of ATL reasoning is that it provides access to some of the most advanced model checking techniques and tools, such as \mcmas~\cite{LomuscioQR:CAV09-MCMAS}, that are in active development within the agent community. 

ATL formulae are built by combining propositional formulas, the usual temporal operators---namely, $\bigcirc$ (``in the next state''), $\Box$ (``always''), $\Diamond$ (``eventually''), and $\U$ (``strict until'')---and a \emph{coalition} \emph{path quantifier} $\coalition{A}$ taking a set of agents $A$ as parameter. 
Intuitively, an ATL formula $\coalition{A}\phi$, where $A$ is a set of agents, holds in an ATL  structure if by suitably choosing their moves, the agents in $A$ \emph{can force $\phi$ true}, no matter how other agents happen to move.
The semantics of ATL is defined in so-called \emph{concurrent game structures} where, at each point, all agents simultaneously choose their moves from a finite set, and the next state deterministically depends on such choices.
%%
% Surprisingly, we show here that by suitably changing the game structure used we are able to synthesize the optimal approximation for a given approximate composition problem, as well as extracting the actual controller generator for such approximation.

% Roughly speaking, a safety-game structure represents a game played by two players, \emph{environment}  and \emph{system}, where, at each turn, the former moves and the latter replies. Moves in any given state are subject to constraints---not every move is allowed. 
% %%
% Intuitively, the system's objective is to always be able to reply to the environment's moves so as to satisfy a given (goal) temporal property, while the environment tries to avoid this. 
% %%
% Technically, the task is to synthesize a winning reply-strategy for the system such that the goal holds in all possible (infinite) ``plays'' that may ensue in the game when the system follows such strategy. A state is ``winning'' if there is a  winning strategy from it.
% %%
% When the goal is a safety one, that is, one of the form ``always $\phi$,'' the game is said to be a \emph{safety game}.
% %%
% For more details on synthesis in game structures we refer the reader to~\cite{Asarin-etal:HS95,Pitterman-etal:JCSS11-GR1Synthesis,DeGiacomoPatrizi:WSFM09}.

In order to reduce a behavior composition problem to an ATL model checking problem, \citea{DeGiacomoFelli:AAMAS10} basically define an ATL structure $\M_{\S,\T}$ with one agent per available and target behavior, and one distinguished agent $\contr$ representing the controller. A state $\tup{b_1,\ldots,b_n,t_s,a,t_d,\ind}$ in such a model encodes the current state $b_i$ of each available behavior, the current state $t_s$ of the target, the current action $a$ being requested by the target, the next target state $t_d$ given the request, and the index of the available behavior to which the last action was delegated to.
The initial states of $\M_{\S,\T}$ encode all possible initial configurations of the composition framework---initial states for all behaviors and a legal initial request. Also, the structure is made to encode all legal evolutions of the composition instance.
The task then involves model checking the special formula $\varphi = \coalition{\contr}\Box(\bigwedge_{i=1,\ldots,n} \state_i \not= \error_i)$ (against structure $\M_{\S,\T}$),\footnote{We note that \cite{DeGiacomoFelli:AAMAS10} deals with final states where the composition execution may stop. For simplicity, we have not dealt with final configurations here, but one can easily accommodate them.} %
 which states that the controller agent has a strategy so that none of the $n$ available behaviors end up in an error state. A behavior arrives to a distinguished ``error''state if it is ever delegated an action that it cannot perform. As a result, the controller agent ought to make sure it always delegates actions in the right way so as to satisfy every potential request, that is, it has to solve the composition problem.
Finally, \citeauthor{DeGiacomoFelli:AAMAS10}~\cite[Definition 2 \& Theorems 3 and 4]{DeGiacomoFelli:AAMAS10} show how to extract a correct controller generator---a structure representing all exact compositions---from the set of \emph{winning states} $[\varphi]_{\M_{\S,\T}}$, namely, all those states $q$ in $\M_{\S,\T}$ such that $q \models \varphi$.  
Intuitively, a winning state for them is one in which the current request is legally honored to some available behavior and \emph{all} corresponding successor states are winning.

Surprisingly, it turns out that one can readily adapt \citeauthor{DeGiacomoFelli:AAMAS10}'s reduction to actually synthesize an optimal approximation for a, possibly non-solvable, \emph{deterministic} composition problem (and to extract the corresponding controller generator).
Though it looks counter-intuitive, the key for this is to \emph{include the target behavior in the coalition} so that the joint-strategy also includes selecting which transition from the actual target may be requested.
In other words, we are instead to model check the following formula against structure $\M_{\S,\T}$:
\[
\tilde{\varphi} = \coalition{\contr,\targetagt}\Box(\bigwedge_{i=1,\ldots,n} \state_i \not= \error_i).
\]

In this case, a winning state in $[\tilde{\varphi}]_{\M_{\S,\T}}$ is one in which the target requests actions such that the  controller can (always) legally honor them to an available behavior, and has \emph{some} corresponding successor winning state. Observe here the implicit existential quantification on the requests, as compared with the universal quantification implied in \citea{DeGiacomoFelli:AAMAS10}'s encoding for exact composition synthesis. 

Intuitively, the idea behind formula $\tilde{\varphi}$, as opposed to formula $\varphi$, is that the coalition is now in control of what can be requested (and what should not be). This suggests that the coalition has the ability to select which parts of the target can be executed without driving the available system into an ``error'' state (due to an impossible fulfilment of a request).
It follows then that one can extract an \emph{optimal} approximation from the \emph{maximal} winning set $[\tilde{\varphi}]_{\M_{\S,\T}}$, as the following result demonstrates.

\begin{theorem}
Let $\S = \tup{\B_1,\ldots,\B_n}$ be a system and $\T = \tup{T,\A,t_0,\varrho_T}$ a target module.
Then, behavior $\hat{\T} = \tup{\hat{T},\A,\hat{t_0},\hat{\varrho}}$ is an optimal approximation for $\T$ on $\S$, where:
\begin{itemize}
  \item $\hat{T} = \set{ \tup{b_1,\ldots,b_n,t_s} \mid  \tup{b_1,\ldots,b_n,t_s,a,t_d,\ind} \in [\tilde{\varphi}]_{\M_{\S,\T}}} \cup \set{\hat{t_0}}$;
%   \item $\A = \set{a| \tup{b_1,\ldots,b_n,t,a,ind} \in W}$ is the finite set of actions;
  \item $\hat{t_0} = \tup{b_{10},\ldots,b_{n0},t_0}$ is the initial state of $\hat{\T}$;
  \item $\hat{\varrho}(\tup{b_1,\ldots,b_n,t_s},a,\tup{b_1',\ldots,b_n',t_d})$ \emph{iff} 
  for some action $a' \in \A$,  and indexes $\ind,\ind' \in  \set{1,\ldots,n}$, it is the case that:
  \begin{itemize}
  \item $\tup{b_1,\ldots,b_n,t_s,a,t_d,\ind},\tup{b_1',\ldots,b_n',t_s',a',t_d',\ind'} \in [\tilde{\varphi}]_{\M_{\S,\T}}$; and
  \item $\tup{b_1,\ldots,b_n,t_s,a,t_d,\ind}$ may transition to $\tup{b_1',\ldots,b_n',t_s',a',t_d',\ind'}$ in $\M_{\S,\T}$.
  \end{itemize} 
\end{itemize} 

\end{theorem}

It is not hard to see that the controller generator~\cite{SardinaPDG:KR08} for $\hat{\T}$ can be extracted by keeping those behavior delegations that transition a winning game state into another winning state in $\M_{\S,\T}$. 
In terms of computational complexity, the model checking task on ATL can be done in polynomial time wrt to the size of the game structure~\cite{AluraHK:JACM02-ATL}. Since the size of such space is exponential on the number of available behaviors, computing the optimal approximation can be done in exponential time (for deterministic systems).
Observe that, in the worst case, the approximation problem itself is (at least) exponential, as it subsumes the classical behavior composition problem (which is known to be EXPTIME-complete even under deterministic behaviors). 
Indeed, in order to check if a problem has an exact composition one can compute its optimal approximation and test (in polynomial time) if it is simulation equivalent  with the original target.

The full details of the ATL encoding, together with an implementation in \mcmas of our running example, can be found in the Appendix.

\section{Discussion}\label{sec:conclusion}

We have proposed a qualitative framework for approximate behavior composition in which the task is to find the \emph{closest} possible target module  that can be implemented with the available modules. 
To that end, we relied on the formal notion of simulation and that of imported controllers for the specification of the problem, and on ATL model checking for actual computation of solutions for the special case of deterministic systems. %(Proofs can be found in~\cite{us:ECAI12}.)
To our knowledge, this is the first account that is able to accommodate behavior composition instances with no complete solutions---arguably the most common ones---while still remaining within the original problem formulation.

% Approximation of transition systems (both discrete and continuous) has been a topic of detailed study; see, e.g., ~\cite{GirardPappas:IEEE2007,chutinan:2001IEEE}. 
% %
%%
Initially, the work of \citea{GirardPappas:IEEE2007} appeared to be extremely related to our objectives, as  it proposes a notion of transition system approximation based on the notion of simulation. However, their work differs in \emph{what} is being approximated.
%%
% Like our approach, \citea{GirardPappas:IEEE2007} proposed a notion of transition system approximation based on the formal notion of simulation. However, their work differs in \emph{what} is being approximated.
%%
In the most general notion of simulation, only some aspects of states are observable and two states in simulation are meant to coincide on their observable aspects. In \citeauthor{GirardPappas:IEEE2007}'s account, an approximate transition system is allowed to differ on such observables up to some extent: $s$ simulates $s'$ implies $s$ can (always) replicate all moves of $s'$ and $s$'s observation is ``similar'' to that of $s'$. 
It follows then that the approximating transition system \emph{must} still be able to mimic \emph{all} actions of the approximated system.
In our framework, there is no notion of state observations (every state has the same observations) and hence we only focus on the   similarities of states in terms of the potential behavior they can generate.
We believe though that one can use their account of approximation when performing  composition \emph{within a shared environment} (as in~\cite{DeGiacomoS:IJCAI07,StroederPagnucco:IJCAI09}), so as to allow the environment to  evolve ``close enough'' to what is necessary.
%%
% We believe though that one can combine our framework with theirs for developing a framework of composition \emph{within a shared environment} (as in~\cite{DeGiacomoS:IJCAI07,StroederPagnucco:IJCAI09}), which is only required to evolve ``close enough'' to what is expected.

Confronted with a behavior composition problem instance admitting no complete solution (i.e., no exact composition) one can, of course, think of other approaches orthogonal to the one developed here.
For example, one may look for additional available behavior modules or enhancement of existing ones with new capabilities that will recover exactness. In some cases, simply adding extra ``copies'' of existing modules could be enough. Thus, installing an extra video camera in the house may turn the problem solvable. 
One could also consider a framework where essential and optional functionalities can be specified, and look for controllers that fully realize the former ones while optimizing the latter ones.
We shall focus on these ideas on future work, as well as on generalizing the actual synthesis techniques from Section~\ref{sec:solution} to nondeterministic systems, possibly relying on more expressive games using GR(1) formulas~\cite{Pitterman-etal:JCSS11-GR1Synthesis}.

The only approach, as far as we know, to deal with unsolvable composition instances is the one we pursued previously in~\cite{YadavSardina:AAMAS11} within a decision-theoretic framework. 
There, the idea is to look for a controller that maximizes the \emph{``expected realizability''} of the target behavior. There are however two major differences with our current proposal. First, their controller may in some runs yield dead-end situations, that is, states from where no further target request can be fulfilled. Under our framework, the user (of the target) can never arrive to those ``error'' situations, as the optimal approximation is always fully implementable.
Second, in our work we kept the strict uncertainty setting from the composition problem found in the literature---no extra knowledge of the domain is assumed to be available. We note that it is well known that \emph{strict} uncertainty cannot always be reduced to a setting where the uncertainty can be measured~\cite{French:DT86}. 
Nonetheless, it would be interesting to be able to accommodate extra domain knowledge \emph{when available}.

\bibliographystyle{abbrvnat}	%% First name abbreviated G. De Giacomo
\bibliography{behcomposition}

  \newpage
  \appendix
  \section{Appendix}
  \newcommand{\bstate}{\text{\emph{state}}}
\newcommand{\sch}{\text{\emph{sch}}}
\newcommand{\req}{\text{\emph{req}}}
\newcommand{\err}{\text{\emph{err}}}
\newcommand{\dATL}{\text{\emph{d}}}
\newcommand{\act}{\text{\emph{act}}}
\newcommand{\start}{\text{\emph{start}}}
\renewcommand{\start}{\sharp}
\newcommand{\ispl}{\textsc{ISPL}\xspace}
%\newcommand{\Post}{\text{\emph{Post}}}
%%%%%%%%%%%%%%%%%%%%%%%%%%%%%%%%%%%%%%%%%%%%%%%%%%%%%%% THEOREM 3 START %%%%%%%%%%%%%%%%%%%%%%%%%%%%%%%%%%%%%%%%%%%%%%%
\subsection{Computing optimal approximations for deterministics behaviors}

Here, we detail the use of ATL model checking technique to compute the optimal target approximation for problem instances involving deterministic available behaviors.
%We detail here the technique to compute the optimal approximation for deterministic systems using ATL model checking.
%%
First,  we show how to construct a concurrent game structure for ATL from a given behavior composition problem. Following that, we present the formula to check in such a model in order to get the optimal approximation. %%

So, let $\S = \tup{\B_1,\ldots,\B_n}$ be a system, with deterministic available behaviors $\B_i = \tup{B_i,\A_i,b_{i0}, \varrho_i}$, for $1 \leq i \leq n$, and let $\T = \tup{T,\A,t_0,\varrho_T}$ be a target behavior.
We start by modifying each available behavior $\B_i$ by adding a new disconnected error state $\err_i$, for each $1 \leq i \leq n$. The error state captures wrong delegations by the controller, i.e., a behavior reaches the error state if it cannot execute the delegated action from its current state.
Let $\Post_i(s,a)$ denote the set of successors states of behavior $\B_i$ after executing action $a$ from its state $s$. Formally, $\Post_i(s,a)=\{s' \mid  \tup{s,a,s'} \in \varrho_i \}$.

We define the \emph{concurrent game structure}, for a system $\S$ and target $\T$, as the tuple $\M_{\S,\T}=\tup{\set{1,\ldots,n,\targetagt,\contr},Q,\Pi,\pi,d,\delta}$, where:
\begin{itemize}
  \item There are $n+2$ players, one per available behavior (agents ${1,\ldots,n}$), one agent for the target module (agent $\targetagt$), and one agent for the controller (agent $\contr$).
  
  \item The states $Q$ of the game structure consists  of the following finite range functions:
  \begin{itemize}
    \item $\bstate_i \in B_i \cup \set{\err_i}$ returns the current state of behavior $\B_i$;
    
	\item $\sch \in \set{i, \ldots, n}$ returns the index of the available behavior that performed the last transition request;
    
    \item $\req \in \varrho_T$ returns the next transition request of the target. Given a transition request $r=\tup{t_s,a,t_d}$, we denote its action $a$ by $\act(r)$.
  \end{itemize}
%   $Q \subseteq B_1 \times \ldots \times B_n \times (T \times \A \times T) \times sch$, where $sch=\{1,\ldots,n \}$ is the set of game states. A game state $\tup{b_1,\ldots,b_n,t_s,a,t_d,ind} \in Q$ captures a system instance, i.e., the available behavior $B_i$ is in state $b_i$ for $1 \leq i \leq n$, the target's next transition request is $t_s\goto{a}t_d$, and $ind$ is the index of the available behavior to which the previous request was delegated.
  \item $\Pi$ is the set of propositions asserting value assignments to the above defined functions.
  \item $\pi$ is the mapping from a game state $q$ to the values returned by the above defined functions. For convenience, we write $\bstate_i(q)=b$ instead of $(\bstate_i=b) \in \pi(q)$.
  \item The function $\dATL(j,q)$ captures the moves available to player $j$ at state $q$, and is defined as follows:
  \begin{itemize}
    \item Available behaviors ($j \in \set{1,\ldots,n}$): 
\[
    \dATL(j,q)  =   
    \begin{cases}
    \{\err_j \}, \text { if } \Post_j(\bstate_j(q),\act(\req(q))=\emptyset 
    \\[2ex]
    \{s \mid s  \in \Post_j(\bstate_j(q),\act(\req(q))\}, \text{ otherwise.} 
    \end{cases}
\]

    \item Target behavior: 
    \[
    \dATL(k-1,q)  = \set{ \tup{t_s,a,t_d}  \in \varrho_T  \mid  \req(q) = \tup{t_s',a',t_s} \text{ for some } t_s',a'  }.
    \]
    
    \item Controller: $\dATL(k,q)= \set{1,\ldots,n}$.
  \end{itemize}
  
  \item $\delta: Q \times \Pi_{i=1}^n (B_i \cup \set{\err_i}) \mapsto Q$ is the game transition function, where $\delta(q,j_1,\ldots,j_k) = q'$ if:
  \begin{itemize}
    \item $\sch(q')=j_k$;
    \item $\state_{i}(q')=j_i$ if $i=j_k$;
    \item $\state_i(q') = \state(q)$ for $i \in \{1,\ldots,n \} \setminus j_k$; and
    \item $\req(q') = j_{k-1}$.
  \end{itemize}
\end{itemize}

We observe that our model is similar to the one used in~\cite{DeGiacomoFelli:AAMAS10} except for the target agent's requests involve transitions rather than actions. 
%\myi the target agent's action consists of a transition request instead of just an action; and \myii the target's moves involve transitions rather than actions. 
%The initial state of the model consists of the following assignments: the behaviors are in their initial states, target is requesting a dummy transition $\start$, and last scheduled behavior ($\sch$) is $\start$. Note, $\start$ is used only for the initial state.

Lastly, we model check the following ATL formula in the structure model $\M_{\S,\T}$:
\[
\tilde{\varphi} = \coalition{\contr,\targetagt}\Box(\bigwedge_{i=1,\ldots,n} \state_i \not= \error_i).
\]
%%
%\notem{CHECK}
In particular, as Theorem~\ref{theo:target-approx} demonstrates, the winning set $[\tilde{\varphi}]_{\M_{\S,\T}}$ provides the basis for building an optimal approximation target.
The code for the implementation of the example in \mcmas can be found at the end of the appendix.

\subsection{Proofs}
\setcounter{theorem}{1}
\begin{theorem}%\label{theo:optaprox_unique}
An optimal approximation $\aprxtgt$ of a target $\T$ on a system $\S$ is unique upto simulation equivalence.

\begin{Proof}
Let $\aprxtgt_i=\tup{T_i,\A,t_{i0},\varrho_i}$ where $i\in \set{1,2}$ be two optimal approximations of $\T$ on $\S$ (wlog we assume $T_1$ and $T_2$ are mutually disjoint).
Let $C_1$ and $C_2$ be exact compositions of $T_1$ and $T_2$ on $\S$, respectively.
Assume $\aprxtgt_1$ and $\aprxtgt_2$ are not simulation equivalent, i.e., $\aprxtgt_1 \not\preceq \aprxtgt_2$ and $\aprxtgt_2 \not\preceq \aprxtgt_1$.
We will show that in such a case $\aprxtgt_1$ and $\aprxtgt_2$ are not optimal approximations of $\T$ on $\S$.
Consider a target behavior $\aprxtgt=\tup{T,\A,t_0,\varrho}$ defined as follows:
\myi $T=T_1 \cup T_2 \setminus \set{t_{10},t_{20}} \cup \set{t_0}$; \myii $\varrho = \varrho'_1 \cup \varrho'_2$, where $\varrho'_i$ is same as $\varrho_i$ except that $t_{i0}$ is replaced by $t_0$ in the transition relations.
See that $\aprxtgt$ is the result of joining $\aprxtgt_1$ and $\aprxtgt_2$ at their initial state, and $\aprxtgt$ simulates both $\aprxtgt_1$ and $\aprxtgt_2$, i.e., $\aprxtgt_1 \prec \aprxtgt$, $\aprxtgt_2 \prec \aprxtgt$.
Since by definition, $\aprxtgt_1 \prec \T$ and $\aprxtgt_2 \prec \T$, and $\aprxtgt$ is union of $\aprxtgt_1$ and $\aprxtgt_2$, it holds that $\aprxtgt \preceq \T$.
Therefore, $\aprxtgt_1 \prec \aprxtgt \preceq \T$ and $\aprxtgt_2 \prec \aprxtgt \preceq \T$.

Next, consider a controller $C$ for $\aprxtgt$ such that it is union of $C_1$ and $C_2$. That is, $C(h,t\goto{a}t') = C_1(h,t\goto{a}t')$, if $\tup{t,{a},t'} \in \varrho'_1$; $C(h,t\goto{a}t') = C_2(h,t\goto{a}t')$, if $\tup{t,{a},t'} \in \varrho'_2$; $C(h,t\goto{a}t') = u$, otherwise. 
Since $C_1$, $C_2$ are exact compositions of $\aprxtgt_1$, $\aprxtgt_2$ on $\S$, respectively, $C$ is an exact composition of $\aprxtgt$ on $\S$.
Therefore, $\aprxtgt$ is an approximation of $\T$ on $\S$. Since $\aprxtgt_1$ and $\aprxtgt_2$ are simulated by $\aprxtgt$, they are not optimal approximations of $\T$ on $\S$.
\end{Proof}
\end{theorem}

\begin{theorem}\label{th:optimality}
Let ${\aprxtgt_1}$ and $\aprxtgt_2$ be two target approximations of target $\T$ on system $\S$, and let $\tilde{C_{1}}$ and $\tilde{C_{2}}$ be exact compositions of $\aprxtgt_1$ and $\aprxtgt_2$, resp.
Suppose also that ${\aprxtgt_2} \preceq {\aprxtgt_1}$ (i.e, ${\aprxtgt_1}$ simulates ${\aprxtgt_2}$).
Then, for every controller $C_1 \in \Omega^{\T}_{\tup{\tilde{C_{1}},\aprxtgt_1}}$, there is no controller $C_2 \in \Omega^{\T}_{\tup{\tilde{C_{2}},\aprxtgt_2}}$ such that $C_2 > C_1$ holds. 

\begin{Proof}
Assume controllers $C_1$ and $C_2$ as above such that $C_2 > C_1$.
Let $\expl_{\aprxtgt_1}$ and $\expl_{\aprxtgt_2}$ be the target explanatory functions that $C_1$ and $C_2$ are built upon, resp.
Now, consider a target explanatory function $\expl_{\aprxtgt_1}'$ for $\aprxtgt_1$ such that $\expl_{\aprxtgt_1}'([h])$ simulates $\expl_{\aprxtgt_2}([h])$ state-wise (i.e., at each step).
%%
% That is, for all histories $h \in \H_{\S}$ if $\expl_2(h) = t_2^0\goto{a^1}\cdots t_2^{|h|}$ then, $\expl_1'(h) = t_1^0\goto{a^1}\cdots t_1^{|h|}$ such that for all $i\leq |h|$ it is the case that $t_2^i \preceq t_1^i$.
%%
Note such function $\expl_{\aprxtgt_1}'$ exists since $\aprxtgt_1$ simulates $\aprxtgt_2$.
Next, consider the imported controller $C_1' \in \Omega^{\T}_{\tup{\tilde{C_1}, {\aprxtgt_1}}}$ built upon target explanatory function $\expl_{\aprxtgt_1}'$. It is not hard to prove that, because traces obtained using $\expl_{\aprxtgt_1}'$ simulate those obtained using $\expl_{\aprxtgt_2}$, $C_1' \geq C_2$ holds (i.e., $C_1'$ dominates $C_2$).
Since, by assumption, $C_2 > C_1$, it follows that $C_1' > C_1$, a contradiction since $C_1$ is not strictly dominated by any induced controller from $\aprxcntrl$ to $\T$.
\end{Proof}
\end{theorem}
%%%%%%%%%%%%%%%%%%%%%%%%%%%%%%%%%%%%%%%%%%%%%%%%%%%%%%% THEOREM 3 END %%%%%%%%%%%%%%%%%%%%%%%%%%%%%%%%%%%%%%%%%%%%%%%

%%%%%%%%%%%%%%%%%%%%%%%%%%%%%%%%%%%%%%%%%%%%%%%%%%%%%%% THEOREM 4 START %%%%%%%%%%%%%%%%%%%%%%%%%%%%%%%%%%%%%%%%%%%%%%%
\begin{theorem}\label{theo:imported_are_max}
% Let $\S$ be an available system and $\T$ a target behavior. 
%%
Let $\aprxtgt$ be an optimal approximation of target $\T$ on system $\S$, and $\tilde{C}$ be an exact composition for $\aprxtgt$.
Then, 
\begin{itemize}
  \item For all $C \in \Omega^{\T}_{\tup{\tilde{C},\aprxtgt}}$, it holds that ${C} \in \Max(\S,\T)$; and
  
  \item $\bigcup_{C \in \Omega^{\T}_{\tup{\tilde{C},\aprxtgt}}} \Delta_{\tup{\S,\T}}^C = \bigcup_{C \in \Max(\S,\T)} \Delta_{\tup{\S,\T}}^C$, that is, all imported controllers account together for all realizable target traces.
\end{itemize}

\begin{Proof}
%The proof uses an auxiliary definition to enhance a behavior to account for a set of traces. Roughly speaking, $\B_{+\Delta}$ is the behavior that results from extending behavior $\B$ with a disjoint sub-transition system that produces exactly those traces (from another behavior) in set $\Delta$.
The proof uses an auxiliary definition to enhance a behavior to account for a set of traces. If $\B= \tup{B,\A,b_0,\varrho}$ is a behavior and $\Delta$ is a set of traces of some other behavior $\B'$ (wlog we assume $\B'$ and $\B$ have disjoint set of states), we define $\B_{+\Delta} =\tup{\hat{B},\hat{\A},b_0,\hat{\varrho}}$ as follows:
\begin{itemize}
  \item $\hat{B} = B \cup \set{ b' \mid b' \text{ is a state in some trace in } \Delta }$;

  \item $\hat{\A} = \A \cup \{ a \mid a \text{ occurs in some trace in } \Delta \}$;

 \item $\hat{\varrho} = \varrho \cup
  \set{ \tup{b_0,a_1,b_1'} \mid b_0' \goto{a_1} b_1' \cdots \in \Delta } \cup
  \set{ \tup{b_i',a_{i+1},b_{i+1}'} \mid b_0' \goto{a_1} b_1' \goto{a_2} \cdots \in \Delta, i\geq 1 }$ .
\end{itemize}

Informally, we extend $\B$ with a disjoint sub-transition system that can produce exactly those traces in $\Delta$. See this is well-defined as $\B'$ is finite, and so will $\B_{+\Delta}$.
For the first claim, we assume there exists $C \in \Omega^{\T}_{\tup{\tilde{C}, \aprxtgt}}$ such that $C \not\in \Max(\S,\T)$.
Hence, there exists a controller $C' \in \Max(\S,\T)$ such that $\Delta^{C}_{\tup{\S,\T}} \subset \Delta^{C'}_{\tup{\S,\T}}$.
We next enhance $\aprxtgt$ with the set of traces realized by $C'$, that is, we build $\aprxtgt_{+  \Delta^{C'}_{\tup{\S,\T}}}$, and extend $\tilde{C}$ to $\tilde{C}'$ such that $\tilde{C}'$ mimics $C'$ for transition requests arising out from $\aprxtgt$'s extension (i.e., requests from traces in $\Delta^{C'}_{\tup{\S,\T}}$).
It can be then shown that $\aprxtgt_{+  \Delta^{C'}_{\tup{\S,\T}}}$ is indeed an approximation of $\T$, and that it has to be simulated by $\aprxtgt$ (or otherwise $\aprxtgt$ would not be optimal approximation).
Because there is a way to evolve $\aprxtgt$ so as to mimic all traces in $\Delta^{C'}_{\tup{\S,\T}}$,  there must exist an induced controller $C^*$ from $\tilde{C}$ into $\T$ such that $ \Delta^{C'}_{\tup{\S,\T}} \subseteq  \Delta^{C^*}_{\tup{\S,\T}}$. This together with the original assumption implies that
$\Delta^{C}_{\tup{\S,\T}} \subset \Delta^{C^*}_{\tup{\S,\T}}$, or what is the same, $C^* > C$, a contradiction since $C$ is an imported controller.

For the second claim, assume there exists a realizable trace $\tau$ of $\T$ such that $\tau$ is \emph{not} realized by any imported controller.
Let $C'$ be a controller realizing $\tau$.
We build the enhanced behavior $\aprxtgt_{+\set{\tau}}$ and extend $\tilde{C}$ to $\tilde{C}'$ so that $\tilde{C}'$ mimics $C'$ for requests arising from $\aprxtgt$'s extension.
Now, $\aprxtgt_{+\{ \tau \}}$ is an approximation of $\T$ and $\aprxtgt$ does not simulate $\aprxtgt_{+\set{\tau}}$ (else $\tau$ would be accounted for by some induced controller), an absurd since $\aprxtgt$ is an optimal approximation.
\end{Proof}
\end{theorem}
\begin{theorem}\label{theo:target-approx}
Let $\S = \tup{\B_1,\ldots,\B_n}$ be a system and $\T = \tup{T,\A,t_0,\varrho_T}$ a target module.
Then, behavior $\hat{\T} = \tup{\hat{T},\A,\hat{t_0},\hat{\varrho}}$ is an optimal approximation for $\T$ on $\S$, where:
\begin{itemize}
  \item $\hat{T} = \set{ \tup{b_1,\ldots,b_n,t_s} \mid  \tup{b_1,\ldots,b_n,t_s,a,t_d,\ind} \in [\tilde{\varphi}]_{\M_{\S,\T}}} \cup \set{\hat{t_0}}$;
%   \item $\A = \set{a| \tup{b_1,\ldots,b_n,t,a,ind} \in W}$ is the finite set of actions;
  \item $\hat{t_0} = \tup{b_{10},\ldots,b_{n0},t_0}$ is the initial state of $\hat{\T}$;
  \item $\hat{\varrho}(\tup{b_1,\ldots,b_n,t_s},a,\tup{b_1',\ldots,b_n',t_d})$ \emph{iff} 
  for some action $a' \in \A$,  and indexes $\ind,\ind' \in  \set{1,\ldots,n}$, it is the case that:
  \begin{itemize}
  \item $\tup{b_1,\ldots,b_n,t_s,a,t_d,\ind},\tup{b_1',\ldots,b_n',t_s',a',t_d',\ind'} \in [\tilde{\varphi}]_{\M_{\S,\T}}$; and
  \item $\delta(\tup{b_1,\ldots,b_n,t_s,a,t_d,\ind},j_1,\ldots,j_{n+2}) = \tup{b_1',\ldots,b_n',t_s',a',t_d',\ind'}$ for some $j_1,\ldots, j_{n+2}$.
  \end{itemize} 
\end{itemize}

\begin{Proof}
Each state $\hat{t}$ of the behavior $\hat{\T}$ is of the form $\tup{b_1,\ldots,b_n,t}$, where $b_1,\ldots,b_n$ are states of behaviors $\B_1,\ldots,\B_n$ and $t$ is a state of the target behavior $\T$; we denote $t$ by $\comp_{\T}(\hat{t})$.
Let $\T = \tup{T,\A,t_0,\varrho_T}$ be the original target behavior.
Due to the definition of $\hat{\varrho}$ in $\hat{\T}$ and $Q$ in the model $\M_{\S,\T}$, it holds that $\hat{t}\goto{a}\hat{t'} \in \hat{\varrho} $ if $\comp_{\T}(\hat{t})\goto{a}\comp_{\T}(\hat{t'}) \in \varrho_T$.  
Now, consider the relation $\R \subseteq \hat{T} \times T$ such that $(\hat{t},t) \in \R$ iff $\comp_T(\hat{t})=t$. Then, for a tuple $(\hat{t},t) \in \R$, for all transitions $\hat{t}\goto{a}\hat{t'}$ in $\hat{\T}$ there exists a transition $t\goto{a}{t'}$ in $\T$ such that $(\hat{t'},t') \in \R$.
See that $\R$ is the simulation relation of $\hat{\T}$ by $\T$,i.e.,  $\hat{\T} \preceq \T$.

Next, we show that $\hat{\T}$ has an exact composition on $\S$. The set $[\tilde{\varphi}]_{\M_{\S,\T}}$ contains all states from where the controller and target can choose their moves so that the behaviors  are never in the error states, i.e., the target can choose which transition to request next such that the controller is able to successfully delegate that transition to a behavior, ensuring realisability of future request(s).
Therefore, for all transitions $\tup{b_1,\ldots,b_n,t}\goto{a}\tup{{b_1'},\ldots,{b_n'},{t'}}$ in $\hat{\T}$ there exists states $\tup{b_1,\ldots,b_n,t,a,{t'},\ind}$,$\tup{b_1',\ldots,b_n',t',a',{t''},\ind'} \in [\tilde{\varphi}]_{\M_{\S,\T}}$ such that the behavior $\B_{\ind'}$ successfully honors the transition request $t\goto{a}{t'}$ and realisation of subsequent transition request ${t'}\goto{a'}{t''}$ can still be guaranteed.
This, in addition with the fact that the initial state of the game is used to initialize the system and the target, is enough to show that $\hat{\T}$ has an exact composition on $\S$. 
%Hence, $\T_W$ is an approximation of $\T$ in $\S$.

Last, we show that $\hat{\T}$ is an optimal approximation of $\T$ on $\S$. 
Let $\aprxtgt=\tup{\tilde{T},\A,\tilde{t_0},\tilde{\varrho}}$ be the optimal approximation of $\T$ in $\S$. Therefore, by definition of optimal approximation, $\hat{\T} \prec \aprxtgt \preceq \T$.
We use proof by contradiction to show that $\hat{\T}$ and $\aprxtgt$ are simulation equivalent.
Assume that $\hat{\T}$ does not simulate $\aprxtgt$, i.e., $\tilde{t_0} \not\prec \hat{t_0}$.
Therefore, there exists a trace $\tilde{\tau}=\tilde{t^{0}}\goto{a^1}\cdots\goto{a^n}\tilde{t^{n}}$ of $\aprxtgt$ such that for all traces 
$\hat{\tau}=\hat{t^0} \goto{a^1} \cdots \goto{a^n} \hat{t^{n}}$ of $\hat{\T}$, there exists a transition $\tilde{t^n}\goto{a^{n+1}}\tilde{t^{n+1}}$ in $\aprxtgt$ for which there is no transition $\hat{t^n}\goto{a^{n+1}} \hat{t^{n+1}}$ in $\hat{\T}$.
That is, $\tilde{\tau}$ cannot be simulated by any trace of $\hat{\T}$.
Let us consider the ATL model $\M_{\S,\aprxtgt}$ between $\aprxtgt$ and $\S$. 
Since $\aprxtgt$ has an exact composition in $\S$, the states $\tilde{W}=[\tilde{\varphi}]_{\M_{\S,\aprxtgt}}$ will accommodate for all $\aprxtgt$'s transition requests. 
Let $\tilde{W}_{\tilde{\tau}} \subseteq \tilde{W}$ be the set of winning states catering for $\tilde{\tau}$'s transitions, that is, $\tup{b_1,\ldots,b_n,\tilde{t_s},a,\tilde{t_d},\ind} \in \tilde{W}_{\tilde{\tau}}$ if $\tilde{t_s}\goto{a}\tilde{t_d}=\tilde{t^i}\goto{a^{i+1}}\tilde{t^{i+1}}$ for some $i\leq n$.
See that the transition $\tilde{t^n}\goto{a^{n+1}}\tilde{t^{n+1}}$ which breaks the simulation of $\aprxtgt$ by $\hat{\T}$ is also included.
Now consider the set of states in the model $\M_{\S,\T}$ defined by: $U =\set{\tup{b_1,\ldots,b_n,t_s,a,t_d,\ind} \mid \tup{b_1,\ldots,b_n,\tilde{t_s},a,\tilde{t_d},\ind} \in \tilde{W}_{\tilde{\tau}}, \tilde{t_s} \preceq t_s, \tilde{t_d} \preceq t_d}$.
That is, the states are similar to the states in $\tilde{W}_{\tilde{\tau}}$ except for the transition requests. The transition requests of $\aprxtgt$ are replaced by the transitions requests from $\T$ such the corresponding states are in simulation.
Note that these states are not only legal states but also included in the set $[\tilde{\varphi}]_{\M_{\S,\T}}$, i.e., $U \subseteq W$: allocation of simulating transitions to same indexes as in the states of $[\tilde{\varphi}]_{\M_{\S,\aprxtgt}}$ will also satisfy the formula in $[\tilde{\varphi}]_{\M_{\S,\T}}$.
%
%allocating these requests to same behaviors similar to the winning states in $G_{\tup{\S,\aprxtgt}}$ will satisfy the safety condition for the game $G_{\tup{\S,\T}}$.
%%
Therefore, $U$ contains states having transition requests $t\goto{a}t'$ of $\T$, corresponding to $\tilde{\tau}$'s transition $\tilde{t^n}\goto{a^{n+1}}\tilde{t^{n+1}}$  such that $\tilde{t^n} \preceq t$ and $\tilde{t^{n+1} }\prec t'$.
%%
%Since the system is same in the games $G_{\tup{\S,\aprxtgt}}$ and $G_{\tup{\S,\T}}$, the behavior evolution will be same for same chosen indexes.
%
%Furthermore, $\aprxtgt$ has an exact solution, so all subsequent requests from the trace $\tau'$ can be successfully delegated. 
%%
%Therefore, all corresponding requests will also be embedded in $W$ - maximal winning set for the game $G_{\tup{\S,\T}}$.
%%
%
Consequently, %for all subsequent $\tilde{\tau}$'s transitions $\tilde{t^n}\goto{a^{n+1}}\tilde{t^{n+1}}$ in $\aprxtgt$ 
there will be a $\hat{\tau}$'s transition $\hat{t^n}\goto{a^{n+1}}\hat{t^{n+1}}$ in $\hat{\T}$ where $\tilde{t^n} \preceq  \comp_T(\hat{t^n})$ and $\tilde{t^{n+1}} \preceq  \comp_T(\hat{t^{n+1}})$, which contradicts the assumption. Therefore, $\hat{\T}$ and $\tilde{\T}$ are simulation equivalent and hence $\hat{\T}$ is an optimal approximation.
\end{Proof}
\end{theorem}

%\notem{CHECK}
See that, if none of the possible ``initial states'' of $\M_{\S,\T}$---where all available and target behaviors are in their initial states and a legal first action is being requested---do not belong to the winning set, then the initial state of the extracted target $\hat{\T}$ (i.e., state $\hat{t_0}$) will end up disconnected from all other states, if any. In that case, it is not hard to see that such approximation will be equivalent to an empty target.

\subsection{Implementation of the house entertainment example}

Below is the code for \mcmas implementation of the house entertainment example presented in the paper. The implementation encodes the given problem in \ispl (Interpreted systems programming language), the input language for \mcmas. \ispl allows defining two different kinds of agents: a number of \emph{standard agents} and an optional \emph{environment agent}. The environment agent offers a common space to share information amongst the standard agents via observable variables (\texttt{Obsvars}). Each \ispl agent definition consists of: \myi set of local states; \myii set of executable actions; \myiii rules to describe which action can be executed in a given state (\texttt{Protocol}); and \myiv an \texttt{Evolution} function describing how the states evolve. Note the similarity between the definition of a \mcmas agent and a behavior module. 

We encode the available behaviors and the target as standard agents and the controller in the environment agent. The environment agent, in particular, has two observables, namely, the currently requested action (\texttt{act}) and the scheduled behavior (\texttt{sch}) to which such action is delegated. Note that the requested action depends on the requested target transition; as evident from the evolution function of the environment. 
The actions for the encoded available behaviors encode their possible evolutions, whereas the actions for the encoded target encode the next possible transition request.
We use the single agent semantics (\texttt{Semantics=SA}) to specify that only one assignment is allowed in each evolution.

We define an evaluation function (\texttt{Error}), evaluated over global states, to capture if any of the available behaviors reaches the error state. A behavior reaches an ``error'' state if it ``skips'' (performs special action ``\texttt{skip}'') when it is actually chosen to be the behavior satisfying the current request. See that a behavior ``skips'' only when all other actions are not possible w.r.t. its protocol. 
Observe also that \mcmas requires the definition of an initial state, from where the system is assumed to begin. In the initial state, all available behaviors and target behavior are in their corresponding initial states, and the action being requested and the scheduled behavior is a dummy action ``\texttt{start}''.  Finally, we define the formula to be model checked: \emph{can the coalition formed by the target and the controller (environment agent in \mcmas) enforce the safety condition of ``not error''?}
{\scriptsize
\begin{verbatim}
Semantics = SA;
Agent Environment
    Obsvars: 
        sch : {GameDevice, MovieDevice,AudioDevice,LightDevice, start};
        act : {movie,game,web,unplug,music,radio,stop, lighton, lightoff, start};
    end Obsvars
    Actions = {GameDevice, MovieDevice,AudioDevice,LightDevice, start};
    Protocol:
        act = start: {start};
        Other: {GameDevice, MovieDevice,AudioDevice,LightDevice};
    end Protocol
    Evolution:
        sch = GameDevice if Action = GameDevice;
        sch = MovieDevice if Action = MovieDevice;
        sch = AudioDevice if Action = AudioDevice;
        sch = LightDevice if Action = LightDevice;
        act = movie if T.Action = t1_movie_t2;
        act = game if T.Action = t2_game_t3;
        act = web if T.Action = t2_web_t3;
        act = music if T.Action = t1_music_t2;
        act = radio if T.Action = t2_radio_t3;
        act = stop if T.Action = t3_stop_t4;
        act = lighton if T.Action = t0_lighton_t1;
        act = lightoff if T.Action = t4_lightoff_t0;
    end Evolution
end Agent

-------------------------------------------------------------
-- GAME DEVICE --
-------------------------------------------------------------
Agent GameDevice
    Vars:
        state: {a0,a1,a2,a3,err};
    end Vars
    Actions = {go_a0,go_a1,go_a2,go_a3,skip};
    Protocol:
        state = a0 and Environment.act = movie : {go_a1};
        state = a1 and Environment.act = game : {go_a2};
        state = a1 and Environment.act = web : {go_a2,go_a3};
        state = a2 and Environment.act = stop: {go_a0};
        state = a3 and Environment.act = unplug: {go_a0};
        Other : {skip};
    end Protocol
    Evolution:
        state = err if Action = skip and Environment.Action=GameDevice;
        state = a0 if Action = go_a0 and Environment.Action=GameDevice;
        state = a1 if Action = go_a1 and Environment.Action=GameDevice;
        state = a2 if Action = go_a2 and Environment.Action=GameDevice;
        state = a3 if Action = go_a3 and Environment.Action=GameDevice;
    end Evolution
end Agent


-------------------------------------------------------------
-- AUDIO DEVICE --
-------------------------------------------------------------
Agent AudioDevice
    Vars:
        state: {b0,b1,b2,err};
    end Vars
    Actions = {go_b0,go_b1,go_b2,skip};
    Protocol:
        state = b0 and Environment.act = music : {go_b1};
        state = b1 and Environment.act = radio : {go_b2};
        state = b2 and Environment.act = stop : {go_b0};
        Other : {skip};
    end Protocol
    Evolution:
        state = err if Action = skip and Environment.Action=AudioDevice;
        state = b0 if Action = go_b0 and Environment.Action=AudioDevice;
        state = b1 if Action = go_b1 and Environment.Action=AudioDevice;
        state = b2 if Action = go_b2 and Environment.Action=AudioDevice;
    end Evolution
end Agent

-------------------------------------------------------------
-- MOVIE DEVICE --
-------------------------------------------------------------
Agent MovieDevice
    Vars:
        state: {c0,c1,c2,err};
    end Vars
    Actions = {go_c0,go_c1,go_c2,skip};
    Protocol:
        state = c0 and Environment.act = movie : {go_c1};
        state = c1 and Environment.act = radio : {go_c2};
        state = c2 and Environment.act = stop : {go_c0};
        Other : {skip};
    end Protocol
    Evolution:
        state = err if Action = skip and Environment.Action=MovieDevice;
        state = c0 if Action = go_c0 and Environment.Action=MovieDevice;
        state = c1 if Action = go_c1 and Environment.Action=MovieDevice;
        state = c2 if Action = go_c2 and Environment.Action=MovieDevice;
    end Evolution
end Agent


-------------------------------------------------------------
-- LIGHT DEVICE --
-------------------------------------------------------------
Agent LightDevice
    Vars:
        state: {d0,d1,err};
    end Vars
    Actions = {go_d0,go_d1,skip};
    Protocol:
        state = d0 and Environment.act = lighton : {go_d1};
        state = d1 and Environment.act = lightoff :{go_d0};
        Other : {skip};
    end Protocol
    Evolution:
        state = err if Action = skip and Environment.Action=LightDevice;
        state = d0 if Action = go_d0 and Environment.Action=LightDevice;
        state = d1 if Action = go_d1 and Environment.Action=LightDevice;
    end Evolution
end Agent


-------------------------------------------------------------
-- TARGET DEVICE --
-------------------------------------------------------------
Agent T
    Vars:
        state: {t0_lighton_t1,t1_movie_t2,t1_music_t2,t2_radio_t3,
                t2_game_t3,t2_web_t3,t3_stop_t4,t4_lightoff_t0};
    end Vars
    Actions = {t0_lighton_t1,t1_movie_t2,t1_music_t2,t2_radio_t3,
               t2_game_t3,t2_web_t3,t3_stop_t4,t4_lightoff_t0};
    Protocol:
        Environment.act = start : {t0_lighton_t1};
        state = t0_lighton_t1 and Environment.act = lighton : 
                                        {t1_movie_t2,t1_music_t2};
        state = t1_movie_t2 and Environment.act = movie : 
                              {t2_game_t3, t2_radio_t3,t2_web_t3};
        state = t1_music_t2 and Environment.act = music : 
                              {t2_game_t3, t2_radio_t3,t2_web_t3};
        state = t2_game_t3 and Environment.act = game : {t3_stop_t4};
        state = t2_radio_t3 and Environment.act = radio : {t3_stop_t4};
        state = t2_web_t3 and Environment.act = web : {t3_stop_t4};
        state = t3_stop_t4 and Environment.act = stop : {t4_lightoff_t0};
        state = t4_lightoff_t0 and Environment.act = lightoff : {t0_lighton_t1};
             
    end Protocol
    Evolution:
        state = t0_lighton_t1 if Action=t0_lighton_t1;
        state = t1_movie_t2 if Action=t1_movie_t2;
        state = t1_music_t2 if Action = t1_music_t2;
        state = t2_radio_t3 if Action = t2_radio_t3;
        state = t2_game_t3 if Action = t2_game_t3;
        state = t2_web_t3 if Action = t2_web_t3;
        state = t3_stop_t4 if Action = t3_stop_t4;
        state = t4_lightoff_t0 if Action = t4_lightoff_t0;
        
    end Evolution
end Agent

Evaluation
    Error if GameDevice.state = err or AudioDevice.state = err or 
             LightDevice.state = err or MovieDevice.state=err;
end Evaluation

InitStates
    GameDevice.state = a0 and AudioDevice.state = b0 and MovieDevice.state = c0
    and LightDevice.state = d0 and T.state = t0_lighton_t1 and 
    Environment.act = start and Environment.sch = start;
end InitStates

Groups
        Coalition = {T, Environment}; -- Approximation
end Groups

Formulae
    <Coalition> G (!Error);
end Formulae

\end{verbatim}
}

\subsubsection{Running result:}

\begin{figure}[ht]
\centering
\includegraphics[scale=.36,angle=-90]{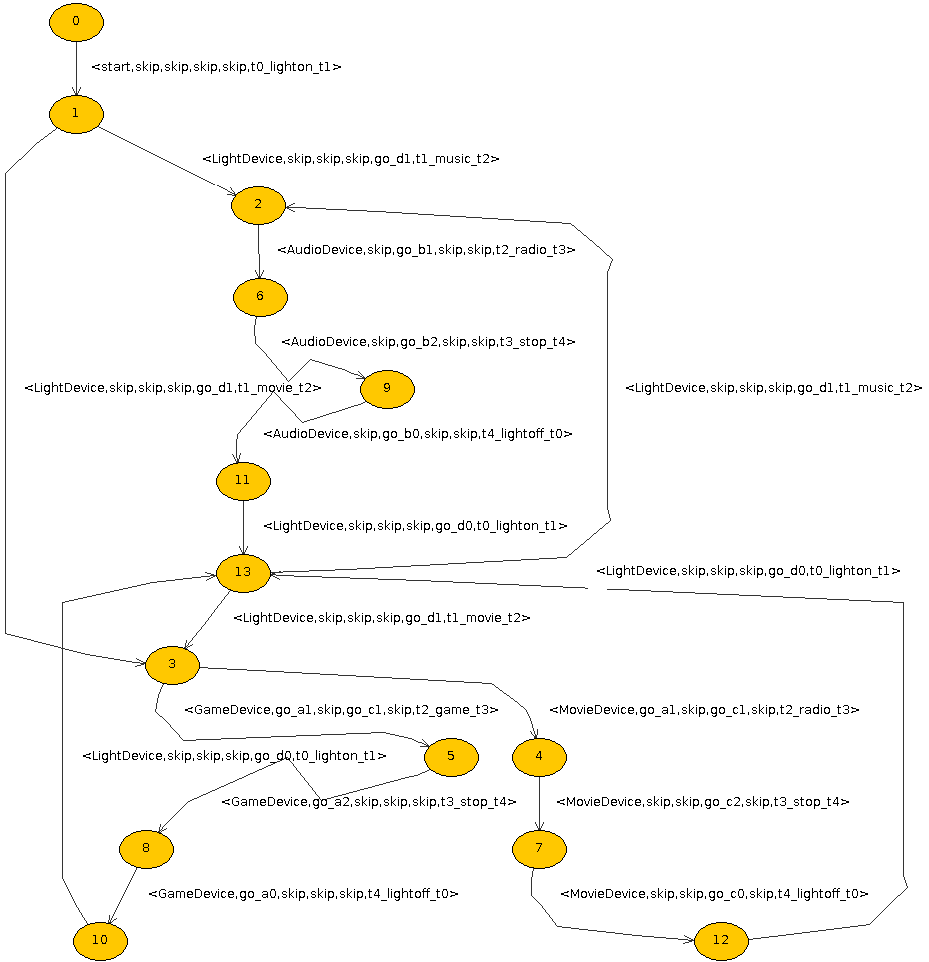}\vspace*{1cm}
\caption{Winning strategy for the House Entertainment example. The transitions in the model correspond to the joint moves of the controller, available behaviors, and the target.} 
\label{fig:model-strategy} 
\end{figure}

Figure ~\ref{fig:model-strategy} shows the witness output by \mcmas for the above translation of the home entertainment example. Extracting the target from this witness, as per Theorem~\ref{theo:target-approx}, yields the optimal target approximation shown in Figure~\ref{fig:ambientIntel-system}.
 
\end{document}